\documentclass[preprint,11pt]{article}
\usepackage{my_symbols}
\usepackage[normalem]{ulem}

\usepackage{array}
\usepackage{times} 

\usepackage{color}			
\newcommand{\alg}[1]{#1} 
\newcommand{\data}[1]{\textsf{#1}}		
\usepackage{caption}
\usepackage{subfig}
\usepackage{blkarray}
\usepackage{url}
\usepackage{rotating} 
\newcommand{\son}[1]{{\begin{sideways}#1\end{sideways}}}

\newcommand{\citep}[1]{\cite{#1}}

\usepackage{col_symbols}

\begin{document}

\title{Multi-label Classification using Labels\\ as Hidden Nodes}

	\author{Jesse Read$^{a,b}$, Jaakko Hollm\'en$^b$}
\date{
	$^a$ DaSciM group. LIX laboratory. {\'E}cole Polytechnique, Palaiseau, France.\\
	$^b$ Dept.\ of Comp.\ Sci., Aalto University and HIIT, Helsinki, Finland.
	}
%




\maketitle

\begin{abstract}



Competitive methods for multi-label classification typically invest in learning labels together. To do so in a beneficial way, analysis of label dependence is often seen as a fundamental step, separate and prior to constructing a classifier. Some methods invest up to hundreds of times more computational effort in building dependency models, than training the final classifier itself. We extend some recent discussion in the literature and provide a deeper analysis, namely, developing the view that label dependence is often introduced by an inadequate base classifier or greedy inference schemes, rather than being inherent to the data or underlying concept. Hence, even an exhaustive analysis of label dependence may not lead to an optimal classification structure. Viewing labels as additional features (a transformation of the input), 
we create neural-network inspired novel methods that remove the emphasis of a prior dependency structure. Our methods have an important advantage particular to multi-label data: they leverage labels to create effective nodes in middle layers, rather than learning these nodes from scratch with iterative gradient-based methods. Results are promising. The methods we propose perform competitively, and also have very important qualities of scalability. 

\end{abstract}



\section{Introduction}
\label{sec:intro}

Multi-label classification is the supervised learning problem where an instance is associated with multiple binary class variables (i.e., \emph{labels}), rather than with a single class, as in traditional classification problems. The typical argument is that, since these labels are often strongly correlated, modeling the dependencies between them allows methods to obtain higher performance than if labels were modelled independently.

As in general classification scenarios, an $n$-th feature vector (instance) is represented as $ \x^{(n)} = [x_1^{(n)},\ldots,x_D^{(n)}]$, where each $x_d \in \R$, $d=1,\ldots,D$. In the traditional \emph{binary} classification task, we are interested in having a model $h$ to provide a prediction for test instances $\x$, i.e., $\yp = h(\x)$. In a multi-label problem, there are $L$ binary output class variables (\emph{labels}), and thus $\ypred = [\yp_1,\ldots,\yp_L] = h(\x)$ outputs a vector of label relevances where $y_j=1$ indicates the relevance\footnote{An equivalent set representation is also common in the literature where, e.g., $\y = [0,1,1] \Leftrightarrow Y = \{\lambda_2,\lambda_3\}$} of the $j$-th label; $j=1,\ldots,L$. 

Probabilistically speaking, $h$ seeks the expectation $\Exp[\y|\x]$ of unknown $p(\y|\x)$. The task is typically posed (\cite{CCAnalysis,MCC2} and references therein) as a MAP estimate,
\begin{equation}
	\label{eq:MAP}
	\ypred = [\yp_1,\ldots,\yp_L] = h(\x) = \argmax_{\y \in \Y} p(\y|\x) 
\end{equation}
where $\Y = \{0,1\}^L$. From $N$ labeled examples (training data) $\D = \{(\x^{(n)},\y^{(n)})\}_{n=1}^N$, we desire to infer predictive model $h$. \Tab{tab:notation} summarizes the main notation used in this work. 

\begin{table}[!hbt]
\small
\begin{center}
\caption{Summary of notation.}
\label{tab:notation}
\begin{tabular}{ll}
\toprule
{\bf Notation } & {\bf Description}  \\
\midrule
	$\x = [x_1,\ \ldots,\ x_D]$ & instance (input vector); $\x \in \mathbb{R}^D$\\ 
$\y = [y_1,\ldots,y_L]$ & $L$-dimensional label/output vector; $\y \in \{0,1\}^L$\\
	$y_j \in \{0,1\}$ & a binary label indicator; $j=1,\ldots,L$ \\
	$\D = \{(\x^{(n)},\y^{(n)})\}_{n=1}^N$ & Training data set, $n=1,\ldots,N$\\
$\ypred = h(\x)$ & predicted label vector given test instance \\
$\yp_j = h_j(\x)$ & individual binary classification for the $j$-th label\\
$\z = [z_1,\ldots,z_K]$ & $K$-dimensional vector of inner-layer variables, $\z \in \{0,1\}^K$\\
\bottomrule
\end{tabular}
\end{center}
\end{table} 

\begin{figure}
	\centering
	\subfloat[][\alg{BR}]{
		\includegraphics[scale=0.95]{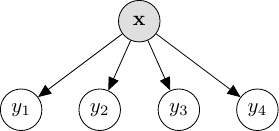}
		\label{fig:br}
	}
	\subfloat[][\alg{CC}]{
		\includegraphics[scale=0.95]{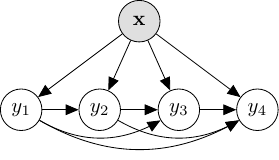}
		\label{fig:cc}
	}
	\caption{\label{fig:graphical} \alg{BR} (\ref{fig:br}) and \alg{CC} (\ref{fig:cc}) as graphical models, $L=4$. For simplicity, we view $\x = [x_1,\ldots,x_D]$ here as single feature attribute. The arrows represent features being used to predict labels (noting that, a predicted label may be reused as a feature attribute value).}
\end{figure}

A well-known and well-used baseline approach is to train $L$ binary models, one for each label. This method is called \emph{binary relevance} (\alg{BR}); illustrated graphically in \Fig{fig:br}. \alg{BR} classifies an $\x$ individually for each of the $L$ labels, as 
\(
	h_{\alg{BR}}(\x) := [h_1(\x),\ldots,h_L(\x)].
\)

Practically the entirety of the multi-label literature points out that the independence assumption among the labels leads to suboptimal performance (e.g., \cite{MBR,MCC2,PCC} and references therein), and that for this reason \alg{BR} cannot achieve optimal performance. A plethora of methods have been motivated by a perceived need to model this dependence and thus improve over \alg{BR}. 
	Many such methods report a performance improvement over \alg{BR}, although often at the cost of a computational trade-off. Despite advances in particular contexts, a `definitive' method for modeling dependence has not yet emerged from the literature.

An early approach is that of Meta-\alg{BR} (\alg{MBR}, also known variously in the literature as `stacked-BR' and `2BR'; \cite{MBR}) which stacks the output predictions of one \alg{BR} model as input into a second (meta) \alg{BR} model, optionally with a skip layer (additionally including the input to both layers), so as to learn to correct the errors of the first model. A related technique is to map infrequent label vector predictions to more frequent label vectors observed in the training data \cite{SM} under some distance function (such as Hamming distance). Error-correcting output coding is a similar schema \cite{MECOC}. 



Rather than `correcting' \alg{BR}'s predictions, several families of methods attempt to learn and classify the labels together. 

The label powerset method (\alg{LP}, see \citep{Overview}) is one such family, where each label vector as a single class value in a multi-class problem. The vector chosen as the `class label' is in fact a binary vector indicating label relevances. It can be seen as directly maximizing \Eq{eq:MAP}, where $\Y = \{\y^{(1)},\ldots,\y^{(N)}\}$ (distinct label combinations in the training data). A number of scalable improvements have been made to this approach (\alg{RAkEL}~\citep{RAkEL2} and \alg{HOMER}~\citep{HOMER} are perhaps two of the most well-known) by dividing up the labelset into a number of smaller more manageable subsets. 

Another example is the family of methods based on classifier chains (\alg{CC}, \cite{ECC2}), illustrated in \Fig{fig:cc}. This model is related to \alg{BR}, but uses binary relevance predictions as extra input attributes for the following classifier, cascaded along a chain. In a probabilistic formulation known as \emph{probabilistic classifier chains} (\alg{PCC}, \cite{PCC}), the method can be derived from  \Eq{eq:MAP} simply by expanding the probability distribution under the chain rule, to obtain
\begin{equation}
	\label{eq:pcc}
	h_{\alg{PCC}}(\x) := \argmax_{\y \in \{0,1\}} p(y_1|\x) \prod_{j=2}^L p(y_j|\x,y_1,\ldots,y_{j-1})
\end{equation}
A Bayes-optimal search of the space will provide an optimal solution for \Eq{eq:MAP}. Such a search is exponential with $L$ and thus typically infeasible, but several approximations exist \citep{MCC2,BeamSearch2,CCAnalysis}. Arguably the most scalable is the greedy search presented in the original paper, scaling linearly with $L$, and allows writing the prediction as
\begin{equation}
	\label{eq:cc}
	h_{\alg{CC}}(\x) := \big[ h_1(\x), h_2(\x,h_1(\x)),\ldots, h_L\big(\x,h_1(\x),\ldots,h_{L-1}(\x,h_1(\x),\ldots)\big)\big]
\end{equation}
where one can note that the prediction of each base model $h_j(\x,\ldots)$ only needs to be evaluated once per test instance (and is then used repeatedly along the chain). Each individual classifier may be phrased probabilistically as 
\begin{equation}
	\label{eq:h}
	\yp_j = h_j(\x,\yp_1,\ldots,\yp_{j-1}) = \argmax_{y_j \in \{0,1\}} p(y_j | \x,\yp_1,\ldots,\yp_{j-1})
\end{equation}
and thereby we simply obtain predictions in order $\yp_1,\ldots,\yp_L$.

In addition to tractable and approximate search methods for inference, a main focus in the development of \alg{CC} methods is the order (and more generally, the structure) of the label nodes in the chain. This is often based heavily around an analysis of `label dependence'. In fact, at least dozens of variations and extensions have appeared in the literature over the past few years (to add to those already mentioned, \citep{BCC,LEAD} are a couple of examples; reviews of many different \alg{CC}-methods is given in \cite{CCAnalysis,MCC2}). They have consistently performed strongly in empirical evaluations, however, the reasons for its high performance are only recently being unravelled. In this paper, we throw new light on the subject.

Two common loss measures used in the multi-label literature are \textit{Hamming loss},
\begin{equation}
	\label{eq:HL}
		L_\textsf{Ham}(\y, h(\x)) = \frac{1}{NL}\sum_{n=1}^{N}\sum_{j=1}^{L}{ \big[ y_{j}^{(n)} \neq \hat{y}_{j}^{(n)} \big] }
\end{equation}
and the subset $0/1$ loss\footnote{Note that we sometimes refer to these in their score form (where higher is better): Hamming score and exact match, respectively}:
\begin{equation}
	\label{eq:EM}
	L_{0/1}(\y, h(\x)) = \frac{1}{N}\sum_{n=1}^{N}{ \big[ \y^{(n)} \neq \ypred^{(n)} \big]}
\end{equation}
where $[a]=1$ if condition $a$ holds, $\ypred=h(\x)$, and $\y = \ypred$ only when both vectors are exactly equivalent. In other words, $0/1$ loss is $1$ wherever Hamming loss is greater than $0$.

From a probabilistic perspective, targeting \Eq{eq:EM} results in \Eq{eq:MAP} (results, e.g., in \cite{OnLabelDependence2,MCC2}). This means that both \alg{CC} and \alg{LP} are suited to $0/1$ loss, but, whereas \alg{CC} methods attempt to make a tractable search of $p(\y|\x)$ using the chain rule, \alg{LP}-based methods typically focus on reducing the possible candidates for $\y$ at inference time (i.e., a much-reduced smaller search space $\Y$). There exist many other variations and extensions for this methodology also, but essentially, discussion often focuses on label dependence and modeling labels together.



In this manuscript we more-thoroughly elaborate on earlier arguments (\Sec{sec:role}), and extend them. In particular we compare different views of label dependence; a probabilistic view, and a neural-network view (\Sec{sec:cc2br}). Building on the latter formulation, we create novel methods (\Sec{sec:novel}). These methods draw inspiration from neural network methods in the sense of having a hidden layer representation consisting of nodes or units but have the particular advantage that they do not need gradient based learners (e.g., back propagation) to learn the hidden nodes. We achieve this with the idea of `synthetic' labels. The resulting methods obtain a high-performing chain without needing to invest effort in chain ordering (contrasting with the earlier methods that invest much computational power into this step). Our evaluation and discussion of methods and their performance is given in \Sec{sec:experiments}. 
Compared to related and competitive methods from the literature, our methods show strong performance on multi-label datasets, and they also provide a strong base for promising future work along the same line. 
Finally, in \Sec{sec:conclusions} we summarize and draw conclusions. 

\section{The Role of Label Dependence in Multi-label Classification}
\label{sec:role}


Throughout the multi-label literature the \emph{binary relevance} method (\alg{BR}) has been cast out of serious consideration on account of assuming independence among label variables. Substantial empirical evidence has been provided indicating that this method can indeed be outperformed by a margin of several, or even tens of percentage points on standard multi-label datasets. 

The main feature of \alg{BR} that sets it aside from more advanced methods is that it models labels as separate problems, and thus ignores the possible presence of dependence among them. Many dozens of papers have been published over recent years presenting new variations of modeling label dependence, and showing that a proposed method improves over \alg{BR}. 

The underlying idea of much published work is that a model structure based on label dependence will lead to a higher-performance than no structure or than a random structure. A typical recipe is first to count co-occurrence frequencies of labels in the training data, and following this, construct a model based on the dependence discovered. The intuition is that, if two labels occur together frequently (statistically more frequently than would be expected at random), then they should be explicitly modeled. \cite{OnLabelDependence2} provides us with an excellent summary from a probabilistic point of view, which has become more common on account of supporting a more formal framework in which to define dependence. This naturally includes the idea that two labels which occur \emph{less} frequently that otherwise expected should also be modeled; implying that human-built topic hierarchies may not necessarily be the best structure (since they tend to separate mutually-exclusive labels into different branches). 
Probabilistically speaking, in the case of marginal \emph{independence}, then
\begin{equation}
	\label{eq:dep}
	p(y_j|y_k) \approx p(y_j) \text{,\quad or equivalently,\quad} p(y_j,y_k) \approx p(y_j)p(y_k)
\end{equation}
for values of some $j$-th and $k$-th labels. This is known as marginal independence, since the input vector ($\x$) is not involved. The distributions $p$ can be approximated by empirical counting in the training set. Given empirical distributions, measures such as mutual information (used in, e.g., \cite{LEAD,MCC2}) can be calculated. Measuring pairwise marginal dependence is fast even for a large number of labels; and thus is a common choice, e.g., \citep{Tenenboim,BCC}.

Models based on marginal dependence may be useful for example for regularizing and `correcting' predictions, as we mentioned in \Sec{sec:intro}, but fundamentally a strong argument exists that models should be based on \emph{conditional} label dependence; since it takes into account the input instance, just as a predictive model does at test time.

 \begin{table}
		 \centering
		\footnotesize
		 \caption{\label{tab:toy}The toy \textsf{Logical} problem: labels correspond to logical operations on the two-bit input. A dataset is created where each combination is sampled with equal probability or not at all (as shown in the final column).}
		\begin{center}
			\begin{tabular}{|llllll|}
			\hline
			             &        & \son{\textsc{or}} & \son{\textsc{and}} & \son{\textsc{xor}} & \\
			 $X_1$       & $X_2$  & $Y_1$             & $Y_2$              & $Y_3$              & $P(\x,\y)$ \\
			\hline
			 0           & 0      & 0                 & 0                  & 0                  & $0.25$ \\
			 0           & 1      & 1                 & 0                  & 1                  & $0.25$ \\
			 1           & 0      & 1                 & 0                  & 1                  & $0.25$ \\
			 1           & 1      & 1                 & 1                  & 0                  & $0.25$ \\
				\ldots   & \ldots & \ldots            & \ldots             & \ldots             & $0$ \\
			\hline
			\end{tabular}
		\end{center}
\end{table}

In \Tab{tab:toy} we introduce a toy example. We may verify (using \Eq{eq:dep} for all values $y_j,y_k \in \{0,1\}^2$) that labels are marginally dependent. For example for $j,k = 2,3$:
		\begin{align*}
			P(Y_2,Y_3) &\neq P(Y_2) \cdot P(Y_3) \\
			[0,0.5,0.25,0.25] &\neq [0.37, 37.5, 0.125, 0.125] 
		\end{align*}
(where $[p_1,\ldots,p_4] = [P(Y_1,Y_2 = 0,0),\ldots,P(Y_1,Y_2 = 1,1)]$ elaborates the distribution). However, in fact in this problem, if we take into account the input, then dependence is absent, for example, if $\x = [0,1]$, then we find that
\[
	P(Y_2|\x,Y_3) = P(Y_2|\x) = 1
\]

Results like this motivate models based on \textit{conditional dependence}. However, obtaining measurements of conditional dependence is particularly intensive, as it inherently involves training classifiers. Indeed, naively $L(L-1)/2$ classifiers should be trained only to obtain pairwise conditional dependence. In \cite{LEAD}, an interesting method was proposed that showed how only $L$ binary models need to be trained, then empirically measure the dependence among the \emph{errors} of the individual models, rather than between labels, as an approximation of measuring conditional dependence. This may not hold rigorously (as explained in \cite{LEAD}) but it is much more efficient. A directed model of nodes is then constructed with off-the-shelf techniques, and a \alg{CC} employed on it.



Variations of this recipe (of measuring dependence then building a structure then training classifiers) have proliferated in the multi-label literature. However, improvements of predictive performance on standard multi-label datasets have begun to reach a plateau, and usually the top recent methods claiming to leverage label dependence in a particular way will invariably not outperform each other by any significant margin (see, e.g., \cite{ExtML}). Consequently, several authors have begun questioning the implied logic that more technique and computational effort invested in discovering and modeling ground-truth label dependence will lead to better predictive performance versus BR and other methods. 

In fact, \cite{OnLabelDependence2} make the case that it should be possible to make risk-minimizing predictions without any particular effort to detect or model label dependence. In other papers (e.g., \cite{ECC2,CCAnalysis}) authors ponder if \alg{BR} has been underrated and could equal the performance of advance methods with enough training data, hence a possible implication that the big data era will render label-dependence models irrelevant. This seems to throw into doubt, or at least under harder scrutiny, the bulk of the contributions to the multi-label literature. 


Indeed, in spite of the widely-perceived need to model label dependence, it can sound almost unusual when phrased in everyday terms. Consider an image-labeling task with two labels (say, \texttt{beach} and \texttt{urban}), and two human volunteers each tasked assigning (or not) one of the labels to a set of images. The assumption that CC will outperform BR is analogous to saying that the labeller of beach scenes performs better if having viewed the decisions of the labeller of urban scenes. We would arguably find this surprising, and perhaps even be tempted to question the ability of the beach-scene classifier. 

This intuition is correct, if we are to focus only on the beach-scene classifier, and to assume that the task of recognising beaches is independent to recognising urban scenes, having observed the scene itself. Formally: to assume conditional independence and evaluation under Hamming loss. Under these assumptions, if the beach-scene classifier performs better after observing the other's predictions, it is correct to assume that he/she might actually not be very good at recognising beaches, in spite of the training data, and is perhaps at least partially guessing relevance based on the fact that the presence of urban scenes more likely than not excludes the presence of a beach scene. 

This illustrates how the choice of base classifier may affect measured label dependence. And it has an important implication: an ideal structure for predictive performance cannot be obtained only from the data, not even large quantities of data, without consideration of the base classifier. The dependence depends on the base classifier, and thus changing the base classifier will also change the dependency model. 

In probabilistic terms, a base classifier may provide us with an approximation of $p(y_j|\x)$ and $p(y_j|\x,y_k)$ (for all label pairs $j,k$, for example), built from the training data. These distributions can be used to measure dependence via \Eq{eq:dep}. But they may not be a particularly good approximation of the true distributions. In \Tab{tab:newtab} we show an example where $p(y_j|\x,y_k) \neq  p(y_j|\x)$ if we use off-the-shelf logistic regression but not if we first pass the input through a basis function expansion. Furthermore, if we speak of CC with the original greedy inference, we must take into account that this was not based on a probabilistic setting, and indeed $h_{\alg{CC}}(\x) \neq h_{\alg{PCC}}(\x)$ (see \Eq{eq:pcc} and \Eq{eq:cc}). 

\begin{table}
		 \centering
		 \caption{\label{tab:newtab}Results of various methods on the \textsf{Logical} problem (see \Tab{tab:toy} and \Fig{fig:newfig}). The base classifier is logistic regression, but \texttt{BR$^{(\phi)}$} first passes the input through a random RBF basis function. The superscripts of \texttt{(P)CC} refer to the models shown in \Fig{fig:log_prob}. \texttt{PCC} refers to optimal inference; and greedy otherwise.}
		\small
	\begin{tabular}{lllllll}
	\hline
		Metric                 & \texttt{BR} & \texttt{CC$^{\subref{fig:log1}}$} & \texttt{CC$^{\subref{fig:log2}}$} & \texttt{PCC}$^{\subref{fig:log2}}$ & \texttt{LP} & \texttt{BR$^{(\phi)}$}   \\
	\hline
		\textsc{Hamming score} & 0.83        & {1.00}                & {0.83}                & {1.00}       & {1.00}      & {1.00}      \\
		\textsc{Exact match}   & 0.47        & {1.00}                & {0.47}                & {1.00}       & {1.00}      & {1.00}      \\
	\hline
	\end{tabular}
\end{table}

Once breaking from a probabilistic graphical framework, the predictive model may lose some connection with the dependence model. This explains why several authors found that to achieve higher accuracy than a random structure, it was necessary to trial a large number of models of different structures via internal validation, and elect the best one (or a top subset) for the final model (as for example in \cite{MCC2}). 
In this manner, there is no separation between finding structure and building a model, as compared other methods that first model dependence, then build a structure based on it. Unfortunately the cost of trialing many models may be prohibitive for large datasets. 

\begin{figure}
	\centering
	\subfloat[][]{
		\includegraphics[scale=1.0]{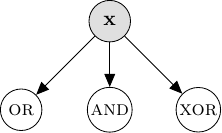}
		\label{fig:log0}
	}
	\subfloat[][]{
		\includegraphics[scale=1.0]{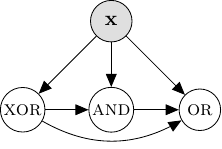}
		\label{fig:log1}
	}
	\subfloat[][]{
		\includegraphics[scale=1.0]{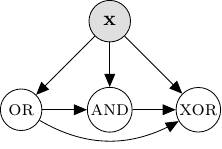}
		\label{fig:log2}
	}
	\caption{\label{fig:log_prob} Models for the \textsf{Logical} toy problem (\Tab{tab:toy}): BR (\ref{fig:log0}) and two initializations of CC (\ref{fig:log1}, and \ref{fig:log2}). Each of these may achieve best accuracy (see also \Tab{tab:newtab}) depending on base learners and inference schemes.}
\end{figure}

Continuing with the running \textsf{Logical} example (\Fig{fig:log_prob}, \Tab{tab:newtab}): Given adequate base models, model \ref{fig:log0} (BR) would be sufficient. However, the linear approximation is inadequate and `causes' dependence -- in the sense that $h_{\textsc{xor}}$ is functionally dependent on at least one of the other base classifiers under CC (with greedy inference) -- and thus only \ref{fig:log2} suffices (from the three models shown in \Fig{fig:log_prob}). This is not true probabilistic dependence, since in that case either \ref{fig:log1} or \ref{fig:log2} represents the joint distribution probabilistically, and performance with an appropriate inference scheme will be equivalent (shown exactly by PCC in \Tab{tab:newtab}). The issue is not one of directionality, since a probabilistic directed model has an equivalent indirected model, already experimented in \cite{CDN} -- and relies on costly inference via Gibbs sampling, thus reviving the original tradeoff faced by \alg{CC} versus \alg{PCC}. The different structures could be trialed individually (as in \cite{MCC2}), but for any reasonable number of labels, this become challenging.  

\begin{figure}
	\centering
	\includegraphics[scale=1.0]{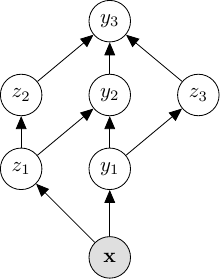}
	\caption{\label{fig:as_nn} Classifier chains as a multi-layer neural network. Units $z_k$ may be viewed as delay units or simply (and equivalently) as \emph{linear} hidden units of weight $1$ such that they do not modify the incoming signal. For example, if weights are defined per layer as $\w_l$ for the $l$-th layer, then $[z_1,\yp_1] = \f_1(\w_1^\top\x)$, $[z_2,y_2,z_3] = \f_2(\w_2^\top[z_1,\yp_1])$ and $y_3 = f_3(\w_3^\top[z_2,y_2,z_3])$. For simplicity, we denote $x = \x$. We assumption that the activation function to $z$-units is linear (and otherwise non-linear). All non-weights are shown as directed arrows, we assume weights from $Z$-units are fixd to $1$, eg., $z_1 = 1 \cdot x = x$. Equivalent recurrent versions with self-loops are also possible.}
\end{figure}

Hence, in consideration of greedy CC, a probabilistic graphical model may give a misleading view.  
In \Fig{fig:as_nn}, we draw CC (i.e., \Eq{eq:cc}) as a neural network, that fires layerwise from $\x$ to $y_3$. Exactly as in CC, $y_3$ takes three input features, in this case $[z_2,y_2,z_3]$ (equivalent to $x,\yp_1,\yp_2$ in CC), and its predictive performance is likely higher than if $y_3$ were estimated directly from $x$ (as \alg{BR} would do). 


To summarize the above: \alg{BR}'s often-cited poor performance is not just a result of failing to detect some ground-truth dependence inherent to the data, but may be due to inadequate base models, which induces dependence. A clean separation between the task of measuring label dependence and constructing a model is often not possible due to approximations made for scalability. In this case, connected models compensate for an inadequacy of the individual base classifiers, by augmenting them with additional features (which may in fact be label predictions). In the following section, we discuss how with adequate base models independence can be maintained across outputs, potentially leading to state-of-the-art results. 



%
%
%
%


\section{Competitive multi-label classification with independent outputs}
\label{sec:cc2br}



In this section we discuss classifiers with independence among the labels. By definition this is a \alg{BR} classifier, except that in the literature it is assumed that \alg{BR} connects inputs \emph{directly} to labels (that there is no hidden layer). We refer to the general case of \emph{unconnected labels} in the sense that there is no direct connection between any of the outputs; the \emph{final} prediction for each individual is made independently of other predictions. 

We motivate classification using such models, and explain the scenarios where such models should perform at least as well as those of fully-structured models. Namely, we take the view that labels can be independent of each other given a sufficiently powerful inner layer. We discuss existing methods that may be viewed as taking this approach, and we lay the framework for our own approach.

\subsection{Motivation for unconnected labels}

In the previous section we explain how classifier chains (\alg{CC}) obtains its performance advantage by using prior labels in the chain as feature-projections for later labels, and that this may compensate for weak or unsuitable base learners (e.g., linear models on non-linear data). Similar conclusions have been separately provided with respect to the \emph{label powerset} method (\alg{LP}) in other work, such as  \cite{OnLabelDependence2}: namely, that predictive power comes as a result of working in a higher space. As mentioned above, \alg{CC} and \alg{LP} represent many of the popular approaches in the multi-label literature, and lie behind many advanced methods.

Although structured models based on CC and LP methods can provide state-of-the-part performance (some examples: \citep{RAkEL2,HOMER,PCC,ECC2}), they nevertheless have some disadvantages: special considerations are necessary to scale up to large numbers of labels, they are often inflexible in the sense of adding/removing labels over time, and have also proved difficult in the past to apply to incremental scenarios such as data streams 
\citep{MEDS2}. A huge number of potential parameterizations and variations of these models requires serious forethought 
and this can be daunting to practitioners, increasing the temptation to go with the `default' option of \alg{BR} and build one separate model for each label. 

Therefore a binary relevance approach, having no inter-linkage among outputs, remains an attractive option. 
This is view is already supported by some of the literature. For example, \cite{LIFT2} reports that having label-specific features for each label results in a very competitive \alg{BR}-type classifier.

Most criticism against \alg{BR} does not involve Hamming loss, since methods predicting labels together (the \alg{BR} and \alg{CC} family; which approximate \Eq{eq:MAP}) target the $0/1$ loss (see \Sec{sec:intro}), whereas Hamming loss is optimized precisely by \alg{BR} (see \cite{OnLabelDependence2}). 
However, there is no general need for a particular preference for $0/1$ loss. Hamming loss is a valid and commonly used metric which often coincides with our intuition in many real-world multi-label tasks -- namely, when we are interested in the accuracy of each label individually. 

\begin{figure}
	\centering
	\subfloat[][BR: $y_{\textsc{or}} = h_1(\x)$]{
		\includegraphics[width=0.35\textwidth]{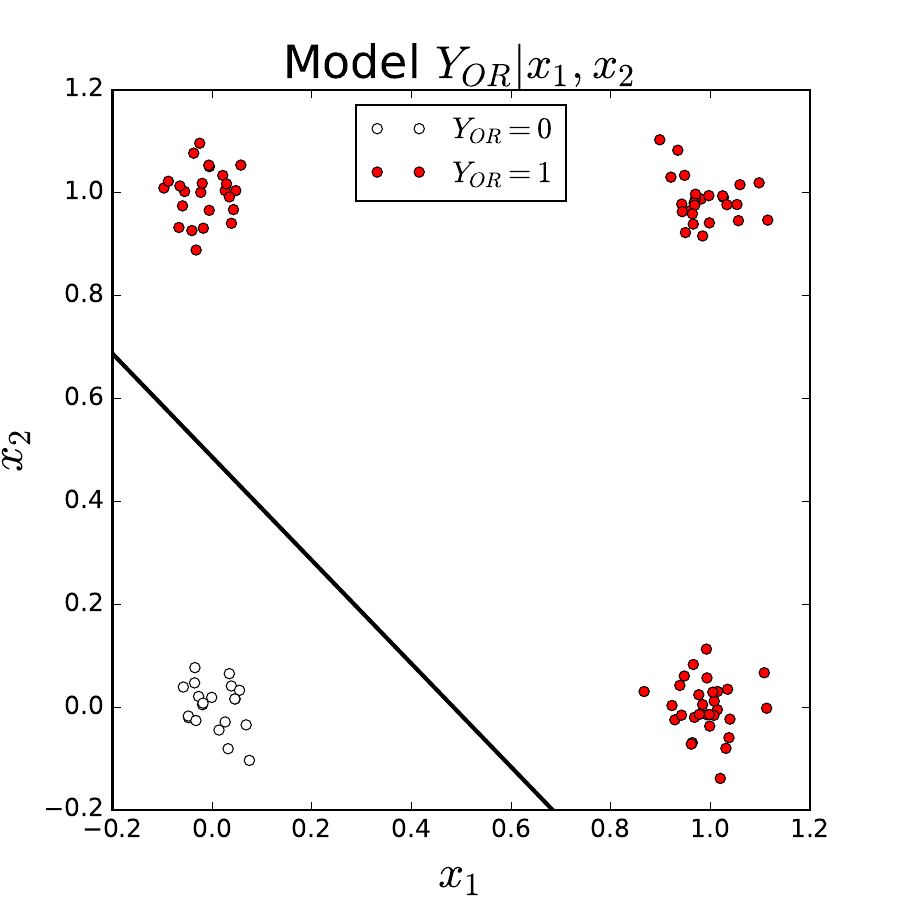}
		}
	\subfloat[][BR: $y_{\textsc{and}} = h_2(\x)$]{
		\includegraphics[width=0.35\textwidth]{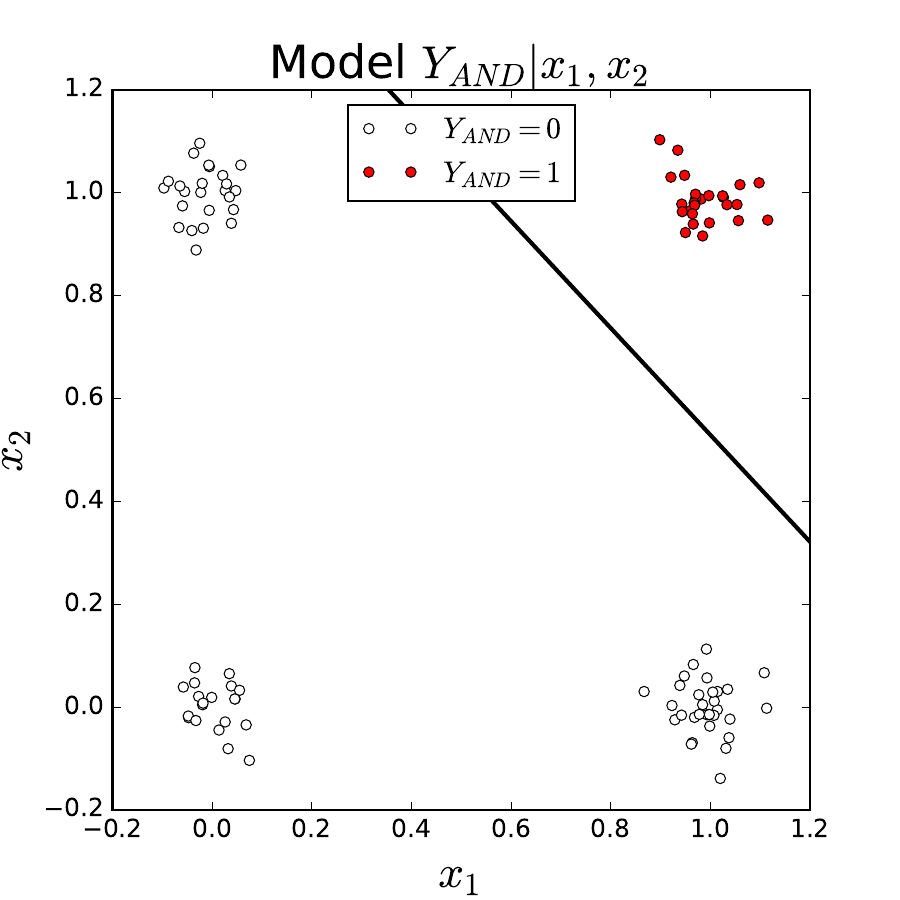}
		}
	\subfloat[][BR: $y_{\textsc{xor}} = h_3(\x)$]{
		\label{f3}
		\includegraphics[width=0.35\textwidth]{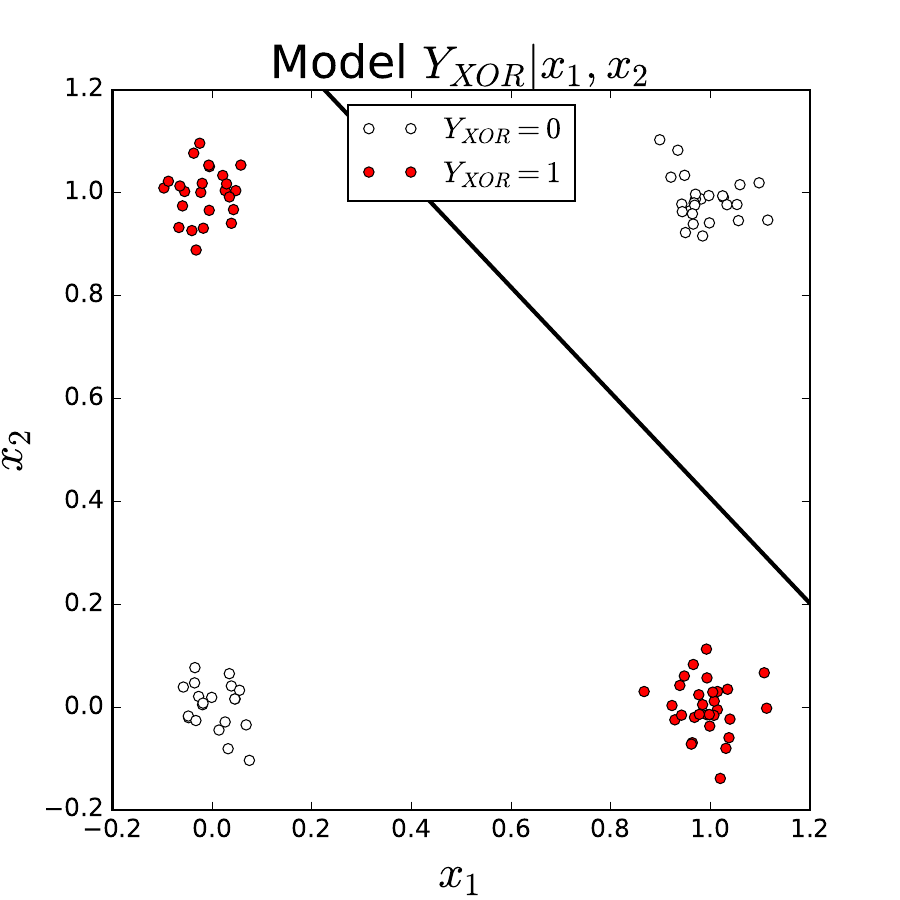}
		}\\
	\subfloat[][CC $y_\textsc{xor} = h_3(\x,\yp_{\textsc{or}})$]{
		\label{f4}
		\includegraphics[width=0.35\textwidth]{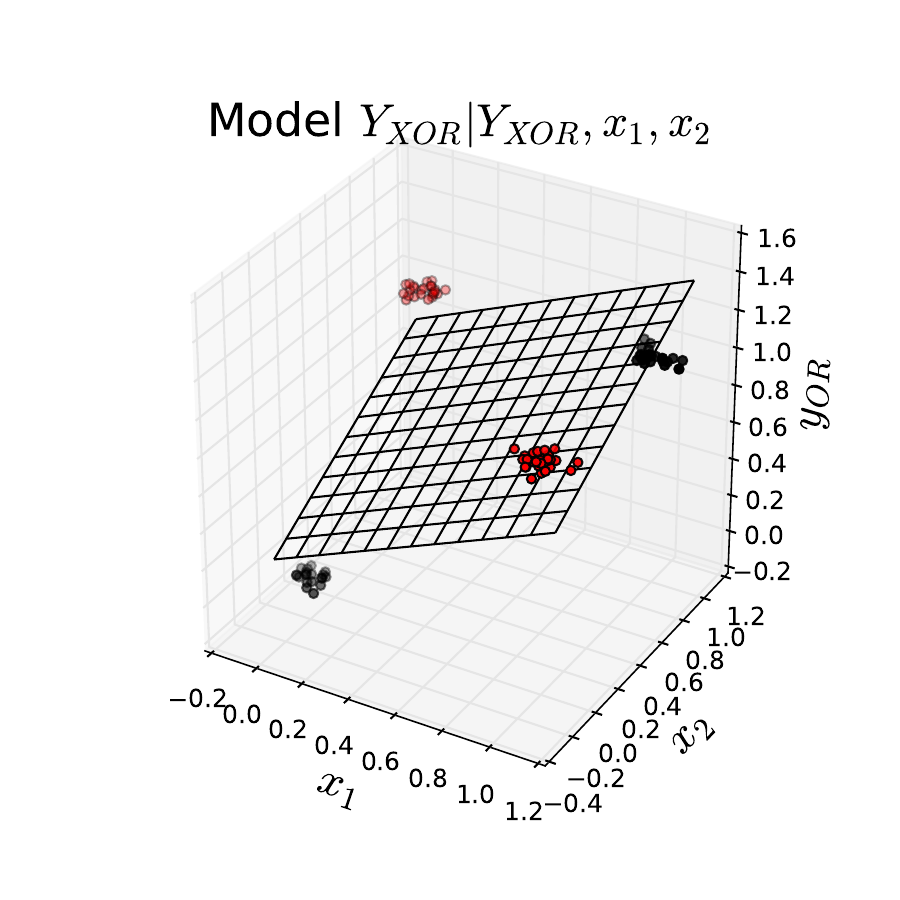}
		}
	\subfloat[][BR $y_\textsc{xor} = h_3(\phi(\x))$]{
		\label{f5}
		\includegraphics[width=0.35\textwidth]{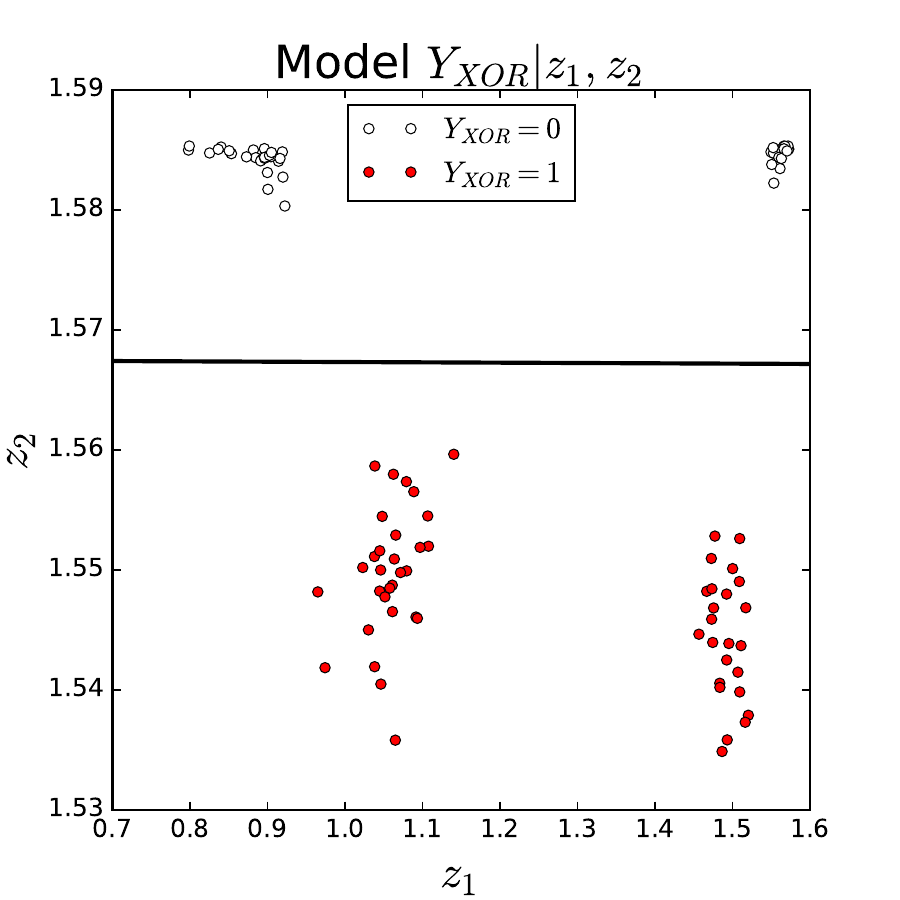}
		}
	\caption{\label{fig:newfig}
	The decision boundaries of base classifiers (logistic regression; shown as black lines/planes) for the three labels of the \textsf{Logical} problem (with some jitter added to the input) of BR (top) and the third label for CC and BR with random basis expansion (in \protect\subref{f4} and \protect\subref{f5}, respectively). Results are given in \Tab{tab:newtab}. For simpler visualization, only one link is considered in \protect\subref{f4}. One can see how $h_3$ fails in \protect\subref{f3} (under BR), but succeeds in \protect\subref{f4} (under CC) and in \protect\subref{f5} (via BR on the new input space). 
	}
\end{figure}

Nevertheless, even under exact match, with \alg{BR} it is empirically possible to outperform structured methods with a suitable base classifier. An example was already illustrated in \Fig{fig:newfig} and \Tab{tab:newtab} (where BR outperforms CC under $0/1$ loss by more than twofold, for particular inference schemes and base classifiers).

\subsection{A framework for multi-label classifiers with structural independence among labels}

\begin{figure}
	\centering
	\includegraphics[scale=1.0]{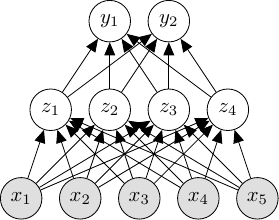}
	\caption{\label{fig:rbm}A multi-label neural network with a hidden layer.}
\end{figure}

In \Fig{fig:as_nn} outputs are independent of each other \emph{in each layer}. $Y_1$ is predicted without knowledge of $Z_1$, and so on. Obviously, $Y_2$ depends directly on $Y_1$, and $Y_3$ depends directly on $Y_2$ and indirectly on $Y_1$ via $Z_3$, and thus predictions are obviously not independent. However, note that from the perspective of a neural network, we may write\footnote{Let us, for the sake of convention, assume that $[y_1,z_1]$ is actually a column vector in the following equation},
\begin{align}
	y_2 &= \w_2^\top[y_1,z_1] \notag \\
	    &= \w_2^\top[h_1(\x),f(\x)] \notag \\
	y_j	&= \w_j^\top\f(\x) = \w_j^\top\z \label{eq:general}
\end{align}
Since $h_1$ and $f$ are both simply functions on the input, we can say that input units $\x$ produce hidden units $\z$ via activation functions $\f = [f_1,f_2]$ (in the case of $j=2$, then $f_1=h_1, f_2=f$) and thus the specific reference to $y_1$ disappears from the equation. Under this generalization we could now draw the framework as \Fig{fig:rbm}. This is a multi-layer network; with independent (in the sense of unconnected) labels, given a fixed $\z$. The layer $\z = [z_1,\ldots,z_H]$ is not actually `hidden' as would normally be the case in such a representation; but can be seen as a feature transformation of $\x$.

\begin{figure}
	\centering
		\includegraphics[scale=0.95]{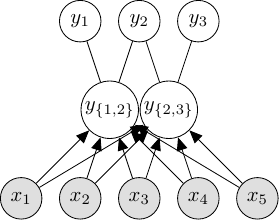}
	\caption{\label{fig:deep_rep}In a \alg{RAkEL}-like scheme, \alg{LP} creates different projections via layers in the output space. Note the comparison to \Fig{fig:as_nn}. We may rename the inner-layer variables to $Z$ to obtain a general representation like that of \Fig{fig:rbm}. }
\end{figure}

We may also arrive at this framework from \alg{LP} methods. In particular, the well-known \alg{RAkEL} method \cite{RAkEL2} extracts $K$ random subsets\footnote{For readers familiar with \cite{RAkEL2} we must emphasise that our notation of $K$ is the number of models/subsets (denoted $m$ in that paper)} of the labelset and applies \alg{LP} to each them; later combining the votes. Let us say that subsets $\{S_1,\ldots,S_K\}$ are chosen where each $S_k \subset \{1,\ldots,L\}$, we may consider inner-layer/meta labels $Y_{S_k}$ (for $k=1,\ldots,K$). Suppose $S_k = \{2,3\}$ (subset of labels $2$ and $3$) which are either both irrelevant or simultaneously relevant to instances the data. Then we could write 
\(
	Y_{S_k} = Y_{\{2,3\}} \in \{[0,0],[1,1]\}
\). 
Assuming that the mapping to the relevant labels is modeled appropriately by the method, we may further simplify notation and simply write 
\(
	z_k \in \{0,1\}
\)
where (in this case) $y_{\{1,2\}} = [0,0] \Leftrightarrow z_1 = 0$, and  $y_{\{1,2\}} = [1,1] \Leftrightarrow z_1 = 1$. Votes for a particular label $y_j$ can be decoded deterministically. See an example of two labels depicted in \Fig{fig:deep_rep}. In \cite{RAkEL2} this is phrased in terms of ensemble voting, but we may generalize this to a sparse vector of weights $\w$ (and bias unit; not shown) and thus write
\begin{equation}
	\label{eq:nn}
	y_j = \w_j^\top\z = \w_j^\top[z_1,\ldots,z_K] = \w_j^\top \f(\x)
\end{equation}
We have recovered again \Eq{eq:general}. Only the formation of units $z_k$ and the composition of weights $\w$ is different between CC and LP. This difference is arguably considerable, but our main point is to bring the two approaches together under a single framework. 

Regarding the performance can we expect from such a framework, this depends on the formation of the inner layer. Empirically, results in the literature show that both greedy-\alg{CC} and \alg{RAkEL}-like methods (as represented by this framework) perform strongly across many metrics. 

As mentioned, \alg{BR} is theoretically the best option when outputs are independent \cite{OnLabelDependence2}. We see that outputs are unconnected in \Fig{fig:rbm} (like \alg{BR}), yet this general framework encompasses two methods that closely approximate \alg{PCC} and \alg{LP}, which in turn optimize $0/1$ loss. 
The reason is in the inner layer, since if outputs are independent of each other given the input, then minimizing Hamming loss and 0/1 loss is equivalent \citep{OnLabelDependence2} (recall, these metrics are described in \Sec{sec:intro}). This is because, it follows that if labels are independent of each other given the input (complete conditional independence), then \Eq{eq:MAP} becomes equivalent to 
\begin{align}
	\ypred 
	    &= \argmax_{\y \in \{0,1\}^L} \prod_{j=1}^L p(y_j|\x) \notag \\
		   &= \argmax_{\y \in \{0,1\}^L} \big[ p(y_1|\x)\cdot p(y_2|\x)\cdot \cdots \cdot  p(y_L|\x) \big] \notag \\
		   &= [\argmax_{y_1 \in \{0,1\}} p(y_1|\x),\ldots, \argmax_{y_L \in \{0,1\}} p(y_L|\x)] \label{eq:bri} 
\end{align}
which is exactly the (probabilistic) formulation of \alg{BR}. This suggests that the power lies in the inner-layer representation. 

Many loss functions can be used in a neural network in a gradient descent scheme\footnote{Hamming loss and $0/1$ loss are interesting in the multi-label context because they are frequently used, minimizers can be found for them, and they can be seen as the two `extremes' of evaluation favoring unstructured and fully-structured evaluation, respectively. There are many other metrics are used in multi-label classification. Among them, $F$-score measures and also particularly the Jaccard index are commonly considered \cite{ECC2,RAkEL2,OnLabelDependence2}. 
A thorough analysis of these other metrics is beyond the scope of this work. 
}. Back propagation will adjust the weights in the inner layer $\z$ to minimize the loss. Unlike Hamming loss, under $0/1$ loss, the error does not decompose across labels (refer to \Eq{eq:HL} and \Eq{eq:EM}, respectively). Observe that, rewriting \Eq{eq:EM}, we achieve
	\begin{align*}
		L_{0/1}(\y, \ypred) &= \big[ \y^{(n)} \neq \ypred^{(n)}\big] \\
							&= \sum_{\y \in \{0,1\}^2} \big[\ypred^{(n)} = \y\big] p(\y|\x) 
	\end{align*}
	Using this to set up a minimization problem, we recover \Eq{eq:MAP}. In this case, we note that only a single error dimension is back propagated and consequently the network treats the output as a single variable \emph{a la} \alg{LP} or \alg{PCC} with full inference, as discussed earlier. 




Yet, despite a non-decomposable loss function, outputs (labels) are still unconnected. They are only connected via input $\z$. The dependence relations are encoded via the inner layer. Hence if using the original input $\x$ does not obtain the required results (high performance given unconnected outputs), we can replace base classifiers or, equivalently, use a powerful $\z$ as input instead, obtained from $\z = \f(\x)$. The very important questions remain of how to build $\f$, and what should $\z$ look like given some $\x$. 
This question has been raised in different points of the literature but has not been definitively answered. Nevertheless, we may at least tackle it. Above we explained how this relates to \alg{CC} and \alg{LP} methods, which we will expand on in \Sec{sec:novel}. Next, we will review some existing tools and techniques for obtaining powerful latent variables (i.e., an inner layer).

\subsection{Existing methods to obtain independence among labels} 
\label{sec:deep_ml}

Although learning latent variable representations is more complex and time consuming than simple input-to-output models, it does provide a means to approach the problem of getting a powerful inner-layer representation. There exists a selection of well-known iterative tools.

One approach to decorrelate labels is to use principle component analysis (PCA) to project from the label space into a new space (of decorrelated labels). This approach was used in, e.g., \citep{PLST, KDE} for multi-label learning. The argument is that if decorrelation is successful, then a \alg{BR} approach is adequate. Predictions into this new space are then simply cast back into the original label space as predictions. One may use the structure of \Fig{fig:rbm}, where $z_k$ is the $k$-th principal component. However, the \emph{linear} decorrelation of PCA is not equivalent to the general concept of obtaining conditional independence. In \cite{KDE} kernels are used to provide non-linearity, but performance is still dependent on the base classifier since conditional dependence may exist; indeed, since PCA is done only on the label space, the input is not considered.

Expectation maximization (EM) is a useful tool to learn many different types of latent variables (i.e., using different distributions) in an iterative fashion, until below some desired threshold bound. A local maximum will be reached.

The popularity of deep learning furnishes the literature with a number of gradient-based techniques which obtain powerful higher-level feature representations of the data in the form of multiple latent layers of variables and non-linearities, rather than a single hidden layer as was exemplified in \Fig{fig:rbm}. These approaches can obtain higher-level abstractions much more more powerful that linear single-layer methods like PCA. Any off-the-shelf method can be used to predict the class label(s) from the top-level hidden layer, or back propagation can be used to fine-tuned all weights for discriminative prediction \citep{DBNbp}.

Despite recent advances in training, e.g., \cite{AdaGrad, Dropout}, deep neural networks usually still require careful configuration and parameter tuning and a lot of data, and as such can rarely be seen as an off-the-shelf out-of-the-box method. Training is computationally expensive due to the large number of iterations required to construct the hidden layers. A key point that we emphasise in this paper is that, in multi-label classification we already have a number of such feature representations available: the labels themselves. As we discussed above, both \alg{CC} and \alg{LP}-based methods can be viewed as as neural networks with inner layers (illustrated in \Fig{fig:as_nn} and \Fig{fig:deep_rep}, respectively). Whether \alg{CC} or \alg{LP} can be thought of as `deep learning' is debatable (the features are not learned in a gradient-based fashion, but rather are given as part of the dataset (as labels) more like basis functions). However, the point we wish to make is this: it is a particular advantage to multi-label data that labels may be treated as additional higher-level features, since these features can be learned or simply created, in a supervised fashion.

Later in \Sec{sec:related} we expand the discussion on other methods that can be used for the task of improving the performance of with nodes.

\section{Using Labels as Hidden Nodes: Novel Methods and Alternatives}
\label{sec:novel}


Building on the material and strategies discussed above, we proposed several novel methods, and compare to existing alternatives. These methods use the label space to construct hidden nodes, i.e., inner layers, that augment the predictive power of the method for estimating label outputs.

\subsection{Classifier chains augmented with synthetic labels (CCASL)}
\label{sec:CCASL}

A common implication of applying \alg{CC} methods is the search for a good order (or structure) for the labels \cite{BeamSearch2,MCC2,OnTheOptimality}. If the label representing a difficult concept is at the beginning of the chain, its base learner may fail to learn it and, even worse, propagate this error to the other learners. This problem cannot easily be identified or solved by a superficial dependency analysis (as we explained in \Sec{sec:role}). 

Since \alg{CC} leverages labels as features in higher-dimensional space, it makes sense that more complex (more abstract, higher level) labels are nearer the end of the chain, where they can be predicted using a more complex input space. However, it is not trivial to assess the complexity of each label, and even if it were, it does not help when all labels are quite complex. Thus, we present a method that adds synthetic labels to the beginning of the chain, and builds up a non-linear representation, which can be leveraged by other classifiers further down the chain. We call this Classifier Chains Augmented with Synthetic Labels (\alg{CCASL}). 

Synthetic labels $z_k|k=1,\ldots,K$ are created as follows: 
\begin{align*}
	\x'_k &= [x_1,\ldots,x_D,z_1,\ldots,z_{k-1}] \\
	\w_k &= [\mB * \W]_{k,1:(D+(k-1))} \\
	z_k &= f(\w_k^\top\x_k') 
\end{align*}
where $\W$ is a $K \times (D + K-1)$ random weight matrix ($W_{j,k} \sim \N(0,0.2)$) and $\mB$ an identically sized masking matrix ($B_{j,k} \sim \textsf{Bernoulli}(0.9)$) such that around 1/10th of weights are zero. A wide selection of activation functions can be selected for $f$. We chose a threshold linear unit (TLU) to make the format of the synthetic labels and true labels both binary, but this is not a requirement. Thus, $z_k = f(a_k) = [a_k > t_k]$ with randomly-selected threshold $t_k \sim \N(\mu_k,\sigma_k \cdot 0.1)$ where $\mu_k$ and $\sigma_k$ is the empirical mean and standard deviation of values $\{a^{(1)}_{k},\ldots,a^{(N)}_{k}\}$ (the equivalent to standardizing or using an appropriate bias on the middle layer). 



\begin{figure}
	\centering
	\includegraphics[scale=1.0]{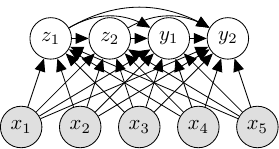}
	\caption{\label{fig:xcc} Classifier Chains Augmented with Synthetic Labels (\alg{CCASL}). Random labels $z_1,\ldots,z_H$ are added to the chain prior to the real labels $y_1,\ldots,y_L$. Any \alg{CC} model can be applied to the top layer.}
\end{figure}

In other words, each value $z_k$ is projected from input $[x_1,\ldots,x_D,z_1,\ldots,z_{k-1}]$, where each one has a slightly more complex space (expanded by one dimension) than the last. \Fig{fig:xcc} shows the resulting network view. It is not an arbitrary choice that the synthetic labels are placed at the beginning: we want to use these labels to improve prediction of the real labels.  
We then train \alg{CC} on label space $\y' = [z_1,\ldots,z_H,y_1,\ldots,y_L]$ and from the predictions ($\ypred'$) we extract the real labels $\ypred = [\yp'_{K+1},\ldots,\yp'_{K+L}] = [\yp_1,\ldots,\yp_L]$ as the final classification. 


As results in \Sec{sec:experiments} will show, this method is able to learn the toy \textsf{Logical} dataset perfectly much more often than \alg{CC}, and does so regardless of the order chosen for the original labels.


We also test \alg{CCASL} on synthetic data, as compared to \alg{BR} and \alg{CC}. We create synthetic datasets using a network of random weights and input, ReLU activation functions on the inner layer and TLUs on the output layer; and also simpler datasets without the hidden layer. Having no hidden layer means that only linear decision boundaries are created. 
	
\Tab{tab:ccVxcc} shows that the difference between \alg{CCASL} (with $K=L$ synthetic labels) and \alg{BR} on the complex synthetic data is significant (nearly 10 percentage points under exact match). The difference between \alg{CC} and \alg{BR} is less significant. Conversely, on the simple linear data, \alg{CC} and \alg{CCASL} both under-perform \alg{BR}. This is not surprising, since the use of a cascade is not needed for the simple data, and this leads to overfitting. 

\begin{table}[t]
	\pink
    \centering 
	\caption{\label{tab:ccVxcc} The average predictive performance of \alg{BR}, \alg{CC} and \alg{CCASL} (with $K=L$) on 150 synthetic datasets of $L=10$ labels, $100$ hidden nodes (only in the case of the complex data) and $N=10000$ examples under 50/50 train/test splits.}
	\subfloat[][\label{tab:sra8}Complex (Non-Linear) Synthetic Data]{
		\small
\begin{tabular}{|l|ll|}
	\hline
	             & Exact Match & Hamm.\ Score \\
	\hline
	\alg{BR}     & 0.558  & 0.843   \\
	\alg{CC}     & 0.573  & 0.843    \\
	\alg{CCASL}  & 0.647  & 0.867    \\
	\hline
\end{tabular}
	}
	\subfloat[][\label{tab:srb8}Simple (Linear) Synthetic Data]{
		\small
\begin{tabular}{|l|ll|}
	\hline
	             & Exact Match & Hamm.\ Score \\
	\hline
	\alg{BR}     & 0.913  & 0.967   \\
	\alg{CC}     & 0.909  & 0.966    \\
	\alg{CCASL}  & 0.893  & 0.959    \\
	\hline
\end{tabular}
	}
\end{table}
 
\begin{figure}
		\centering\pink
		\subfloat[][\label{fig:sra1}
		\begin{tabular}{ccc}
			\hline
			\alg{BR}	& \alg{CC}$^{(1,2)}$/\alg{CC}$^{(2,1)}$ & \alg{CCASL} \\
			\hline
			0.95	& 0.95/0.94 & 0.92 \\
			\hline
		\end{tabular}
		\vspace{0.5cm}
		\\
		\centering
		\begin{tabular}{ll}
			\hline
			Base Classif. & Acc. \\
			\hline
			$h_1(\x)$ (of BR)     & 0.97 \\
			$h_2(\x)$ (of BR)     & 0.97 \\
			$h_1(\x,y_2)$ (of CC) & 0.97 \\
			$h_2(\x,y_1)$ (of CC) & 0.96 \\
			\hline
		\end{tabular}
	]{
		\includegraphics[scale=0.30]{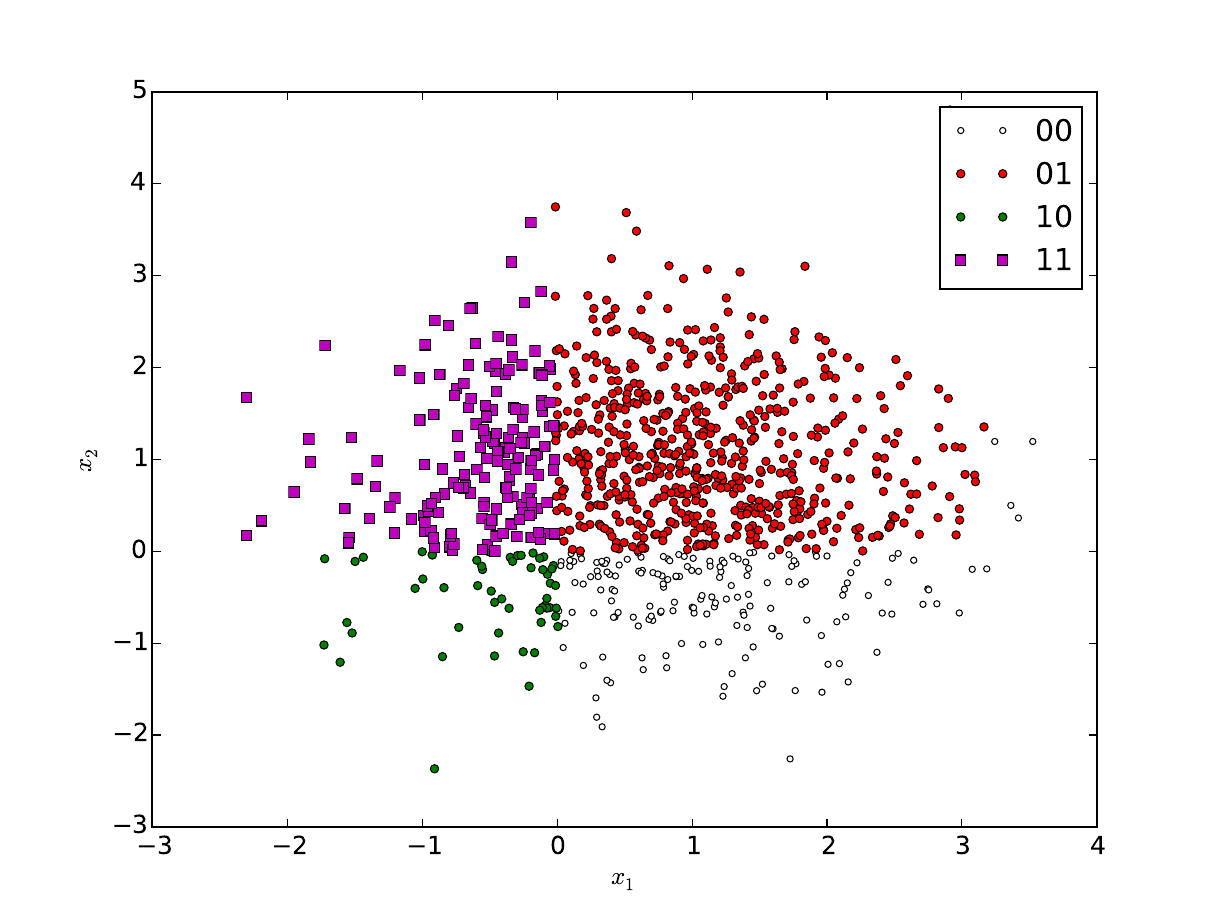}
	}
	\subfloat[][\label{fig:sra2}
		\begin{tabular}{ccc}
			\hline
			\alg{BR}	& \alg{CC}$^{(1,2)}$/\alg{CC}$^{(2,1)}$ & \alg{CCASL} \\
			\hline
			0.58	& 0.53/0.61 & 0.66 \\
			\hline
		\end{tabular}
		\vspace{0.5cm}
		\\
		\centering
		\begin{tabular}{ll}
			\hline
			Base Classif.      & Acc. \\
			\hline
			$h_1(\x)$ (of BR)     & 0.63 \\
			$h_2(\x)$ (of BR)     & 0.31 \\
			$h_1(\x,y_2)$ (of CC) & 0.73 \\
			$h_2(\x,y_1)$ (of CC) & 0.33 \\
			\hline
		\end{tabular}
	]{
		\includegraphics[scale=0.30]{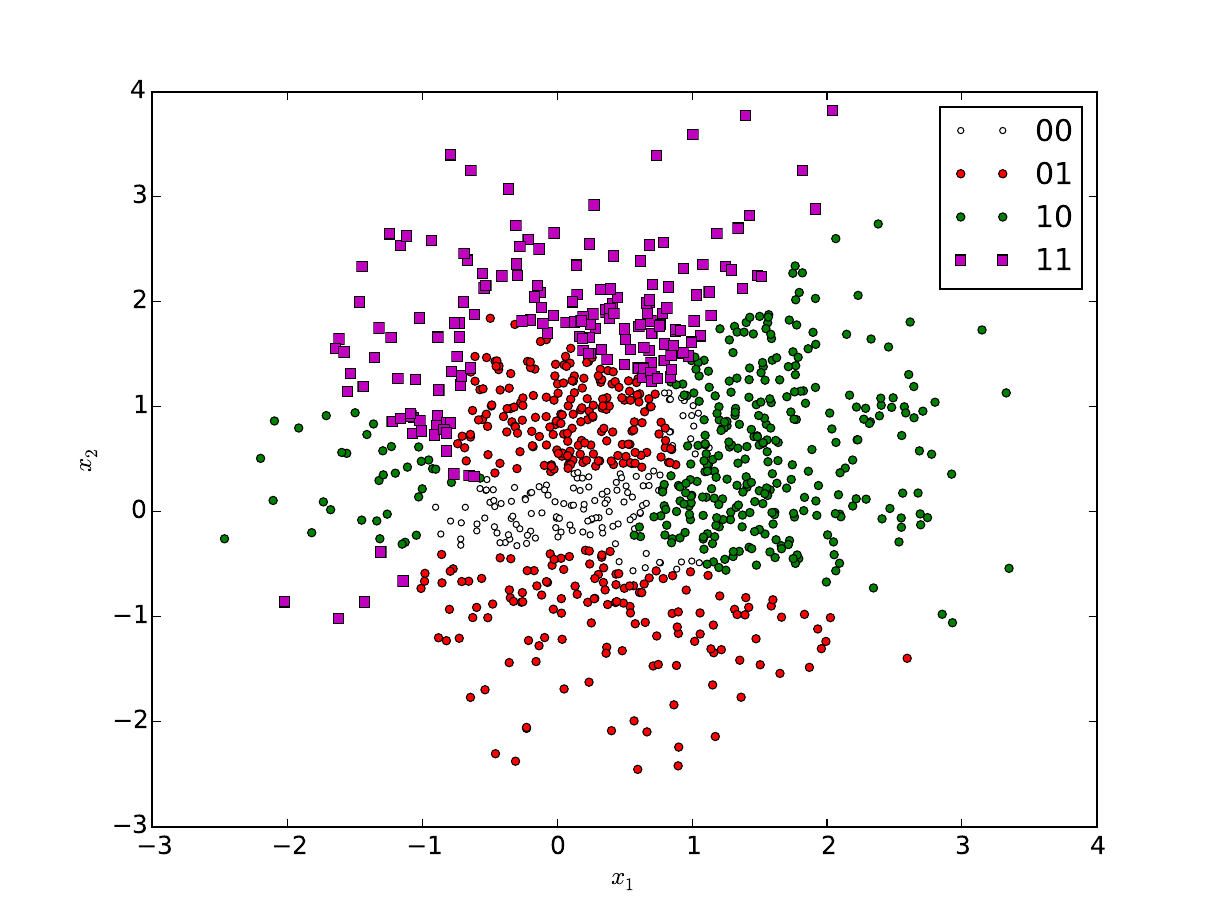}
	}

	\subfloat[][\label{fig:sra8}Simple Synthetic Data]{
		\includegraphics[scale=0.30]{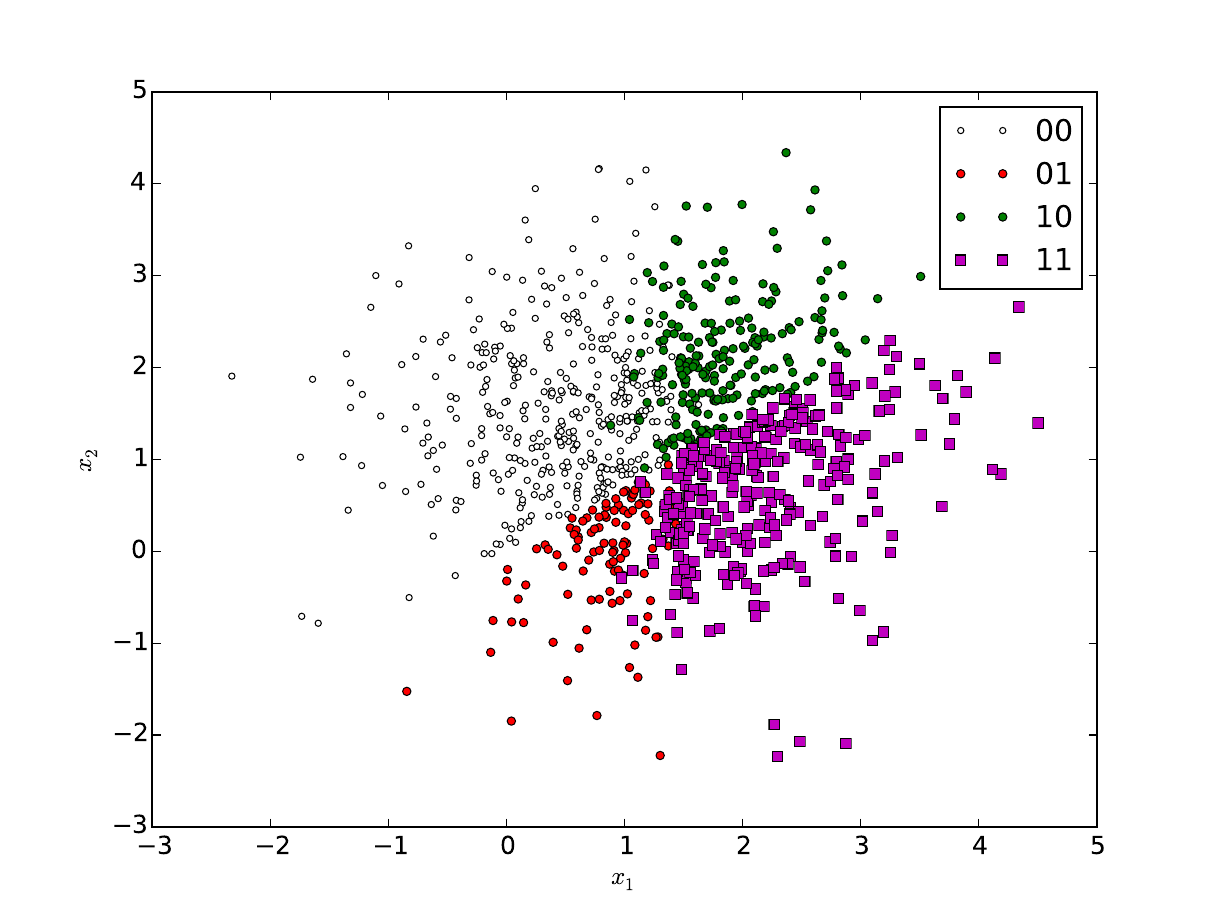}
	}
	\subfloat[][\label{fig:srb8}Simple Synthetic Data]{
		\includegraphics[scale=0.30]{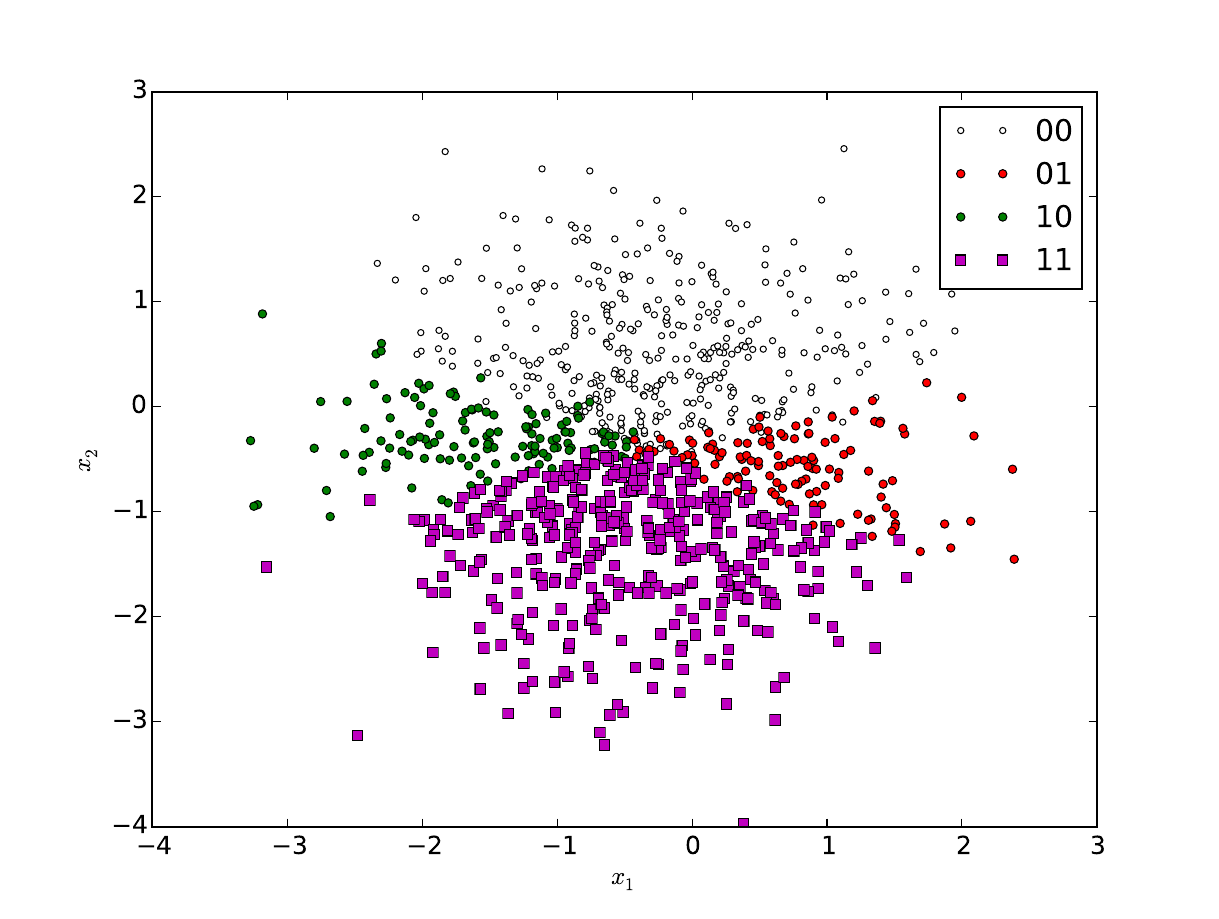}
	}
		\caption{\label{tab:results_syn1} \pink Illustrative results on synthetic data of dimensions $D=2$, $10$ hidden units (for the complex data), and $L=2$ labels. The legend in the plots indicates $y_1y_2$. For the complex data generated with a hidden layer (above) we have provided exact match performance of \alg{BR}, \alg{CC} (with both possible chain orderings), and \alg{CCASL}, and also the accuracy of the base classifiers within \alg{BR} and \alg{CC}. For the simple data (Figs.\ \protect\subref{fig:sra8} and \protect\subref{fig:srb8}) we already know that no dependence exists and thus did not tabulate a similar accuracy analysis. 
		}
\end{figure}

\Fig{tab:results_syn1} displays particular results on smaller dimensions (to visualize easily in a two-dimensional plot). Again we see the contrast in exact match (the results under \Fig{fig:sra1} and \Fig{fig:sra2}). \alg{CCASL} performs relatively much better on the complex data. The accuracy of \alg{BR}'s base classifiers versus those of \alg{CC} (in different label orders) sheds some light on this. Under both orders, \alg{CC} obtains some benefit for the second label, only one of the labels can benefit at a time. Except under \alg{CCASL}, where both may benefit. Conversely, on the simple synthetic data (two samples of which are given in \Fig{fig:sra8} and \Fig{fig:srb8}), \alg{CCASL} displays no advantage at all over \alg{BR}, and in fact does worse, exactly as shown in \Tab{tab:ccVxcc}. 

Essentially \alg{CCASL} creates a cascaded projection of the input space and then combines this space with the label space and learns it together in the style of classifier chains.
Scalability is easily tunable by using a smaller and less-cascaded projection (reducing the number of connections). Indeed, it has already been shown that a sparsely connected \alg{CC} incurrs little to no performance degradation with respect to \alg{CC} in many cases \citep{BCC,MCC2}. But already, relatively few hidden labels are necessary for \alg{CCASL}; in empirical evaluation we obtain good performance with $K=L$.

\subsection{Adding a meta layer: CCASL+BR}
\label{sec:CCASL+BR}

\begin{figure}
	\centering
	\includegraphics[scale=1.0]{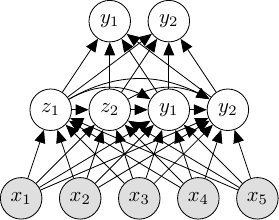}
	\caption{\label{fig:dxcc} \alg{CCASL+BR}: \alg{CCASL} as a middle layer. Outputs are stacked again into a second meta \alg{BR}-classifier (i.e., similarly to \alg{MBR}). By modeling dependence in the middle layer, dependence can be ignored in the output layer.}
\end{figure}

In spite of solving the \textsf{Logical} problem and excelling on synthetic data, our empirical results (in \Sec{sec:experiments}) show that \alg{CCASL} does not appear to out-perform \alg{CC} by much on real datasets. Besides this issue, we already stated a desire for models that have an unconnected label layer. To obtain the independent outer layer and also counteract overfitting, we also add an extra \alg{BR}-layer (i.e., a meta-\alg{BR}) to get \alg{CCASL+BR}. 
\Fig{fig:dxcc} displays the resulting network.


\alg{CCASL+BR} combines the advantages of several important methods in the multi-label literature: namely \alg{BR} (outputs are not directly connected to each other), \alg{MBR} (a form of regularization), and \alg{CC} (a chain cascade to add predictive power). 
Back propagation is not necessary. 

A potentially interesting alternative to the issues faced by \alg{CCASL} is to reconsider a more thorough probabilistic inference, to  avoid error propagation (itself a kind of overfitting). We leave a detailed investigation to future work, but do include an experimentation of this approach in \Sec{sec:experiments}.

\subsection{Adding an augmented layer of meta labels: \alg{CCASL+AML}}
\label{sec:CAML}

In addition to creating synthetic labels from the feature space, it is also possible to create synthetic labels from the label space, as we outlined in \Sec{sec:cc2br}. Namely, we create binary synthetic labels, based on subsets $S_k \subset \{1,\ldots,L\}$. For each such subset, the most common combination $\y^*_{S_k}$ is selected from the training data. This allows us to create binary variables 
\[
	z_k = \big[ \y_{S_k} = \y_{S_k}^* \big]
\]
where, as used above, the notation $\big[a\big]$ returns $1$ if condition $a$ holds (else $0$).

To re-exemplify, if we select $S_k = \{1,3,6\}$, and if we find that the most commonly occurring combination of these labels in the training set is $\y_{S_k}^* = \y^*_{\{1,3,6\}} = [1,0,1]$, then $z_k = 1$ indicates the relevance of this combination, and $1/K$ votes are added to $y_1$, $y_3$, and $y_6$. 

\begin{figure}
	\centering
	\includegraphics[scale=1.0]{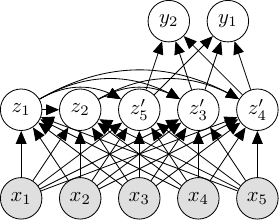}
	\caption{\label{fig:dcc} \pink \alg{CCASL} with an Augmented Middle Layer (\alg{CCASL+AML}). The prime distinguishes the meta labels $z'_k$ from the synthetic cascaded labels $z_k$ (in this illustration).}
\end{figure}

For even more predictive power, we combine these units with synthetic labels generated from the feature space, as does \alg{CCASL}, and hence denote this method \alg{CCASL+AML} (\alg{CCASL} with an Augmented Meta/middle Layer). The resulting network is exemplified in \Fig{fig:dcc}. In the figure, we distinguish the AML units as $z'_k$ from the CCASL units as $z_k$.  

\subsection{Related and alternatives models}
\label{sec:related}



Adding non-linearity to an otherwise linear method via basis expansions is a fundamental technique in statistical learning \citep{EoSL}. Under this methodology, basis functions can be either chosen suitably by a domain expert, or simply chosen arbitrarily to achieve a more dimensioned representation of the input, and higher predictive performance. In this latter case, polynomials are a common choice, such that $\z = x^2,x^3,\ldots,x^p$ up to some degree $p$. Under this approach learning can proceed ordinarily by treating $\z$ as the input. 

Restricted Boltzmann machines (RBMs) are a probabilistic method for finding higher-level feature representations $\z = f(\W^\top \x)$, where $\W$ is learned with gradient-based methods such as to minimize energy $E(\x,\y,\z) = \exp\{-\z^\top\W\x\}$ and $f$ is some non-linearity (typically a sigmoid function $f(a) = \frac{1}{1+e^{-a}}$, but more recently ReLUs have been particularly popular). The RBM can then cast any input instance stochastically into a different-dimensioned space, as $\z$. Any classifier (e.g., BR) can be trained on this space. The RBMs are often stacked on each other greedily and fine tuned for discriminative classification with back propagation \citep{DBNbp} but can also be trained directly for competitive discriminative prediction \citep{CRBM}.

%


The multi-layer multi-label neural-network `BPMLL' proposed earlier by \cite{BPMLL} has become well known in the multi-label literature. Its structure is essentially that exemplified in \Fig{fig:rbm}, and the hidden layer is trained using back propagation. The only difference from a vanilla multi-layer perceptron (MLP) is that it is trained to minimize a rank loss. However, its out-of-the-box performance has consistently been shown to under-compete with the state-of-the-art. Later, a multi-label radial-basis-function (RBF) network was proposed by \cite{MLRBF}. However, this network was not picked up widely in the literature. The limited reception of these methods is possibly due to the fiddly `hit-and-miss' nature of the gradient-based learning (with respect to the selection of learning rates, hidden nodes and so on) which we have specifically tried to avoid with our methods. 

Recently, \cite{NN2} readdressed the application of neural networks for multi-label classification, in particular for the task of large scale text classification. They implemented MLPs with a number of recently proposed techniques from deep learning, namely ReLUs, AdaGrad, and dropout. They also used the standard cross entropy loss function and sigmoid activations functions, as opposed to BPMLL's choice of rank loss. At least on text datasets (the particular domain they tackled) these networks were shown to be highly competitive compared to BPMLL.

Even with the latest techniques, the primary disadvantage often associated with multi-layer neural networks is that of back propagation, which is widely used for discriminative learning, involving many iterations of propagating the input up through the network to the outputs and then the error back down to adjust the weights. This is the case even for stacked deep networks with RBMs for pre-training weights. Indeed, for this reason \cite{NN2} build a case for only using no more than a single hidden layer (despite using the latest deep-learning techniques) on their large datasets. 

An early alternative to back propagation in neural networks was proposed by \cite{CMAC}, to use the idea of random functions to project the input layer into a new space in the hidden layer, rather than learn this hidden layer iterative gradient descent. More recently, this idea has appeared in so-called `extreme learning machines' (ELMs, \cite{ELM}), which are essentially neural networks with random weights. Again, this is built on the premise of eliminating the need for gradient-based methods in the hidden layer to obtain huge speed ups in learning. Radial basis functions are a typical choice in ELMs, making them closely related to (or in fact, an instance of) RBF networks (e.g., \cite{MLRBF}). ELMs are related to the methods we propose. However, unlike \alg{CCASL+AML}, ELMs do not work down from the label layer. Furthermore, in our approach we use a cascade of functions which are treated as synthetic labels and learned in a supervised manner along with the real labels.

Some methods transform the label space, and predict into this space, thus being related to our methods. Already (in \Sec{sec:deep_ml}) we mentioned principal label-space transformation \citep{PLST} and variations \citep{KDE} that make the independent targets, apply independent models to them, and then cast the predictions back into the original label space. The various methods used (such as PCA) are related to RBMs via factor analysis (RBMs can be seen as a binary version thereof). 

Many methods for obtaining latent variables can be cast in a probabilistic graphical model framework (for example RBMs). Already in \cite{CCAnalysis}, parallels are drawn between probabilistic \alg{CC} and conditional random fields (CRFs), which are a general class of latent-variable model. CRFs have been studied with application to multi-label classification by \cite{CollectiveMC}. But as discussed earlier, greedy CC is more comparable to explicit basis function and loses some of its probabilistic connections and is difficult to cast into the indirected nature of a CRF.

Finally, simply using a powerful non-linear base classifier may remove the need for transformations of the feature space altogether. For example, decision trees are non-linear classifiers that have proven themselves in a huge variety of domains, including in multi-label classification, as shown in \citep{ExtML}. 

\section{Experiments}
\label{sec:experiments}

%

We conducted an empirical evaluation on a variety of multi-label datasets commonly used in the literature, comparing to established baseline and related methods from the literature. The main goals are to compare relative performance of our proposed methods under different dataset/metric combinations and to investigate the mechanisms behind this performance. 

\subsection{Setup and methodology}

We compare the novel methods that we developed (namely, \alg{CCASL} and variants) with a selection of methods from the literature: baseline \alg{BR}, \alg{CC} both with greedy inference, and with Monte-Carlo search for inference (the latter we refer to as PCC\footnote{Because it is an approximation of full inference used in \cite{PCC}. See \cite{MCC2} for details, but note also that other efficient PCC methods exist, e.g., beam search \cite{BeamSearch2} or $\epsilon$-approximate \cite{CCAnalysis}}), an ELM, and a MLP. \Tab{tab:methods} summarizes methods and their parameters. References and descriptions can be found in \Sec{sec:intro} and \Sec{sec:related}.  For base classifiers, we used logistic regression in all cases (including ELMs and MLP in the sense that the final layer of these networks uses sigmoid activation functions), \emph{except} when denoted with a superscript RF which denotes random forest as a base classifier (e.g., \alg{BR$_{RF}$}). All methods were implemented in Python\footnote{Code will be made available at \url{https://github.com/jmread/molearn}}, using the \textsc{Scikit-Learn} library\footnote{\url{http://scikit-learn.org}} for logistic regression and random forest.

\begin{table}[t]
	\pink
    \centering 
	\footnotesize
	\caption{\label{tab:methods}\pink Summary of methods. We propose the ones based on \alg{CCASL}. }
\begin{tabular}{lll}
	\toprule
	Method &	Strategy & Parameters \\
	\midrule
	\alg{BR} & Independent outputs & \\
	\alg{CC} &  Cascade across labels & \\
	\alg{PCC} &  \ldots with Monte-Carlo inference & $M=50$ iterations/example \\
	\alg{CCASL} &  CC with synthetic labels & $K=L$ \\
	\alg{CCASL+MBR} &  \ldots as input to a \alg{BR}  & $K=L$ \\
	\alg{CCASL+AML} &  \ldots with additional meta labels  & $K=L, K'=2L^\dagger, |S_k|=3$ \\
	\alg{ELM} &  Random projection to \alg{BR}  & $K=2L$, TLU activation\\
	\alg{BR$_{RF}$} & \alg{BR} with random forest & \\
	\alg{MLP} &  Neural network with back prop.        & $K \in \{2L,2D\}^\ddagger, \lambda=0.1, I=1000$, \\
	          &                                                & sigmoid activation \\
	\bottomrule
\end{tabular}
	\begin{tabular}{rl}
		$^\dagger$ & $K'$ is the number of AML units; each based on random subset $S_k \subseteq \{1,\ldots,L\}$ \\
			$^\ddagger$ & Best of these (on \emph{internal} 60/40 split), but limited to maximum $K=100$ \\
	\end{tabular}
\end{table}





We used 10 datasets typically used in multi-label evaluations\footnote{Available from \url{https://sourceforge.net/projects/meka/files/Datasets/}}, plus a version of the illustrative logical dataset from \Sec{sec:cc2br}. All are listed in \Tab{tab:datasets}.

\begin{table}[t]
    \centering 
    \caption{\label{tab:datasets}A collection of datasets and associated statistics, where LC is \emph{label cardinality}: the average number of labels relevant to each example.}
	\footnotesize
	\begin{tabular}{rrrrcl}
	\toprule
	    		& $N$  &$L$ &	$D$		& LC &Type 		 \\           
	\midrule           
	\data{Logical}       &20    &3     &2      &1.50          &logical \\    
	\data{Music}       &593    &6     &72      &1.87          &audio   \\    
	\data{Scene}   	&2407	&6     &294		&1.07          &image	 \\     
	\data{Yeast}   	&2417 	&14   	&103	   	&4.24          &biology \\     
	\data{Medical}		&978 	&45      &1449    &1.25          &medical/text	 \\     
	\data{Enron}   	&1702	&53    &1001    &3.38          &text	 \\     
	\data{Reuters} 	&6000	&103   &500		&1.46          &text	 \\     
	\data{OHSUMED}		&13929  &23     &1002    &1.66 &text	 \\     
    \data{MediaMill}	&43907  &101    &120		&4.38	       &video	 \\     
	\data{Bibtex}		&7395   &159    &1836    &2.40          &text	 \\     
	\data{Corel5k} & 5000 & 374 & 499 & 3.52 &image \\
	\bottomrule
	\end{tabular}\\
\end{table}

We use the score metrics (higher values are better): Hamming score (the score equivalent of Hamming loss; \Eq{eq:HL}), and exact match (the score equivalent of $0/1$ loss; \Eq{eq:EM}), already detailed in this paper and used extensively in multi-label evaluations. Hamming score rewards methods for predicting individual labels well, whereas exact match rewards a higher proportion of instances with \emph{all} label relevances correct. 

\begin{figure}
	\centering
	\includegraphics[width=0.6\textwidth]{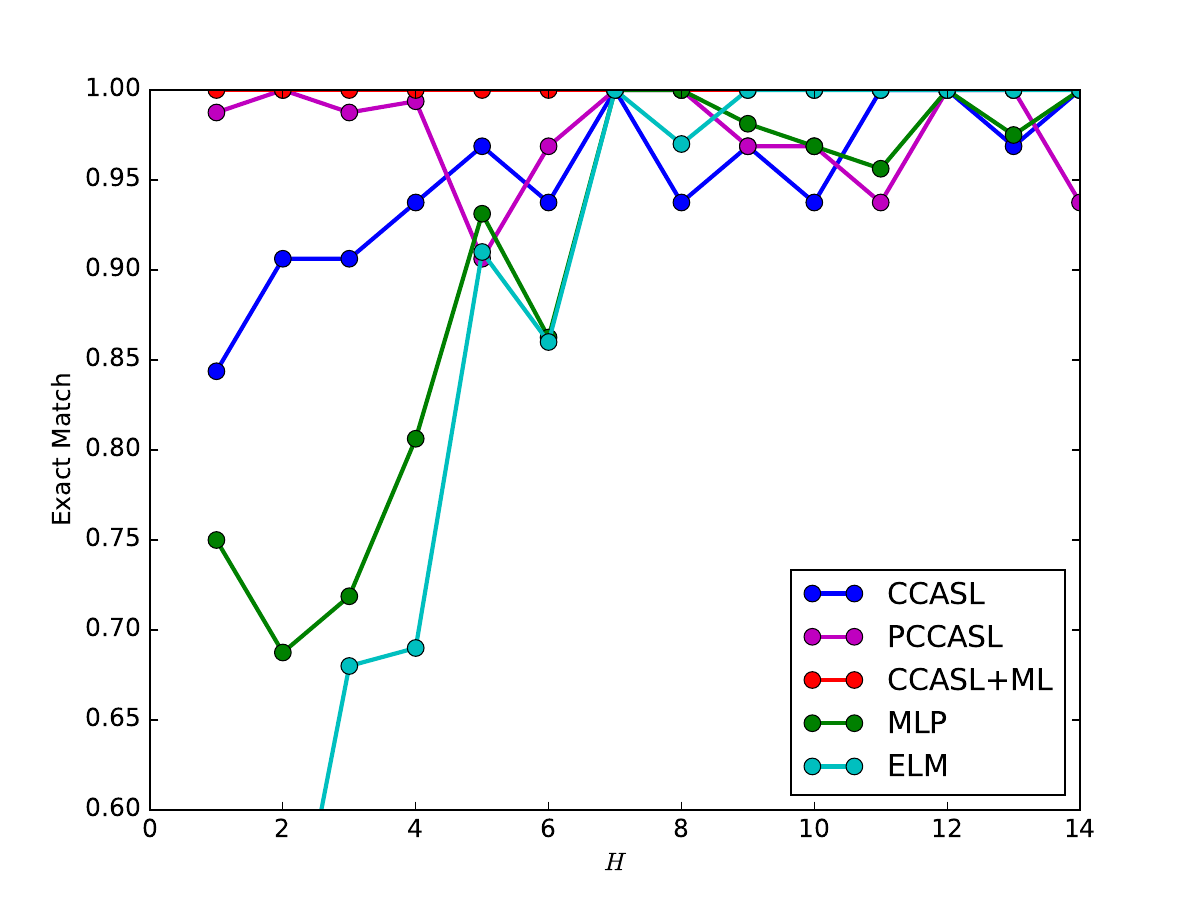}
	\caption{\label{fig:contrast} Different methods compared on the \textsf{Logical} dataset for varying numbers of hidden nodes. Each point represents the average of 10 runs on 60/40 train/test split with randomly shuffled labels. Note that PCCASL refers to CCASL with probabilistic Monte-Carlo inference (as used in PCC in the other experiments); 10 iterations/example in this case.}
\end{figure}

We carry out 10 iterations for each dataset, each time with a random 60/40 train-test split and a random order of labels. \Fig{fig:contrast} shows the effect of different numbers of synthetic labels / hidden nodes for performance on the \textsf{Logical} dataset. \Fig{fig:labup} shows the effect of more or fewer output labels given an equivalent input space. \Tab{tab:results} shows the main results, of predictive performance across all datasets. \Tab{tab:scalability} shows the results in terms of running time. All experiments were run on Intel 2.6 GHz processors. 

\begin{table}
	\pink
	\centering
	\caption{\label{tab:results}Predictive performance in terms of exact match and Hamming score. Average result across $10$ randomizations (of instance and label space) and a 60/40 train/test splits. The average rank is shown after the performance for each dataset. Value rounded to 2dp and rank rounded to 1dp for display only. \alg{CCASL+} is abbreviated to \alg{C.+}.}
	\footnotesize
	\subfloat[][\textsc{Exact Match}]{
	\begin{tabular}{lrrrrrrrrr}
\hline
Dataset   & \alg{BR} & \alg{CC} & \alg{PCC} & \alg{CCASL} & \alg{C.+BR} & \alg{C.+AML} & \alg{ELM} & \alg{BR$_{RF}$} & \alg{MLP} \\
\hline
Logical    & 0.52 9 & 0.64 8 & 0.87 4 & 0.78 7 & 1.00 2 & 1.00 2 & 0.79 6 & 1.00 2 & 0.83 5  \\ 
Music      & 0.23 8 & 0.25 6 & 0.26 2 & 0.25 4 & 0.26 3 & 0.27 1 & 0.17 9 & 0.25 7 & 0.25 4  \\ 
Scene      & 0.47 8 & 0.55 5 & 0.56 3 & 0.54 6 & 0.56 4 & 0.58 1 & 0.21 9 & 0.48 7 & 0.56 2  \\ 
Yeast      & 0.14 6 & 0.18 2 & 0.19 1 & 0.18 3 & 0.17 5 & 0.18 4 & 0.11 8 & 0.10 9 & 0.12 7  \\ 
Medical    & 0.45 7 & 0.46 6 & 0.42 8 & 0.68 3 & 0.70 1 & 0.69 2 & 0.28 9 & 0.68 4 & 0.62 5  \\ 
Enron      & 0.11 7 & 0.12 5 & 0.12 4 & 0.13 2 & 0.13 2 & 0.13 1 & 0.06 9 & 0.12 6 & 0.09 8  \\ 
Reuters    & 0.45 7 & 0.47 4 & 0.47 1 & 0.47 6 & 0.47 2 & 0.47 2 & 0.37 9 & 0.47 5 & 0.38 8  \\ 
Ohsumed    & 0.15 4 & 0.15 3 & 0.15 4 & 0.15 7 & 0.15 7 & 0.15 6 & 0.03 9 & 0.17 2 & 0.21 1  \\ 
M.Mill     & 0.09 8 & 0.12 3 & 0.12 4 & 0.13 1 & 0.11 6 & 0.11 5 & 0.10 7 & 0.12 2 & 0.05 9  \\ 
Bibtex     & 0.10 5 & 0.11 4 & 0.10 6 & 0.17 1 & 0.16 3 & 0.17 2 & 0.04 9 & 0.10 7 & 0.07 8  \\ 
Corel5k    & 0.01 7 & 0.01 4 & 0.00 9 & 0.02 3 & 0.02 1 & 0.02 1 & 0.01 7 & 0.01 5 & 0.01 7  \\ 
\hline
avg rank   &  6.95      &  4.55      &  4.23      &  4.05      &  3.45      &  2.55      &  8.27      &  5.09      &  5.86      \\
\hline
\end{tabular}

		\label{fig:r2}
	}\\
	\subfloat[][\textsc{Hamming Score}]{
	\begin{tabular}{lrrrrrrrrr}
\hline
Dataset   & \alg{BR} & \alg{CC} & \alg{PCC} & \alg{CCASL} & \alg{C.+BR} & \alg{C.+AML} & \alg{ELM} & \alg{BR$_{RF}$} & \alg{MLP} \\
\hline
Logical    & 0.84 8 & 0.88 7 & 0.96 4 & 0.93 6 & 1.00 2 & 1.00 2 & 0.93 5 & 1.00 2 & 0.06 9  \\ 
Music      & 0.79 2 & 0.78 7 & 0.78 6 & 0.78 7 & 0.78 5 & 0.78 4 & 0.75 9 & 0.79 1 & 0.78 3  \\ 
Scene      & 0.87 8 & 0.88 6 & 0.88 4 & 0.88 7 & 0.88 4 & 0.88 3 & 0.84 9 & 0.90 1 & 0.89 2  \\ 
Yeast      & 0.79 1 & 0.78 6 & 0.78 8 & 0.78 7 & 0.78 4 & 0.78 5 & 0.79 3 & 0.79 2 & 0.22 9  \\ 
Medical    & 0.70 9 & 0.72 7 & 0.72 8 & 0.95 6 & 0.98 2 & 0.98 3 & 0.96 5 & 0.99 1 & 0.98 4  \\ 
Enron      & 0.90 8 & 0.92 6 & 0.91 7 & 0.94 3 & 0.94 3 & 0.94 2 & 0.93 5 & 0.95 1 & 0.07 9  \\ 
Reuters    & 0.98 9 & 0.98 7 & 0.98 7 & 0.99 4 & 0.99 2 & 0.99 2 & 0.98 5 & 0.99 1 & 0.98 5  \\ 
Ohsumed    & 0.92 4 & 0.92 5 & 0.92 6 & 0.92 8 & 0.92 8 & 0.92 8 & 0.93 3 & 0.94 1 & 0.93 2  \\ 
M.Mill     & 0.97 2 & 0.97 6 & 0.97 8 & 0.97 6 & 0.97 4 & 0.97 4 & 0.97 4 & 0.97 1 & 0.96 9  \\ 
Bibtex     & 0.98 6 & 0.98 5 & 0.98 7 & 0.99 2 & 0.99 2 & 0.99 2 & 0.97 9 & 0.99 2 & 0.98 7  \\ 
Corel5k    & 0.98 7 & 0.98 5 & 0.97 9 & 0.98 4 & 0.99 2 & 0.99 2 & 0.98 5 & 0.99 1 & 0.98 8  \\ 
\hline
avg rank   &  5.82      &  6.27      &  6.86      &  5.64      &  3.68      &  3.50      &  5.73      &  1.32      &  6.18      \\
\hline
\end{tabular}

		\label{fig:r3}
	}
\end{table}


\subsection{Discussion}

Results highlight the performance of \alg{CCASL} methods. All out-perform baseline \alg{BR} and \alg{CC} overall, particularly with meta layers (\alg{+BR} and \alg{+AML}). Under exact match, the \alg{CCASL+} methods are almost invariably the top performers. There are some exceptions -- under \textsf{Yeast} and \textsf{Ohsumed} -- with respect to \alg{P/CC}. Already it has been observed that that advanced methods typically do not make any significant improvements over each other or \alg{BR} on the \textsf{Yeast} dataset (see, for example, the results of \citep{ExtML}) and indeed we note that under Hamming score, it is \alg{BR} that obtains the top score. On \textsf{Ohsumed} the difference is not distinguishable even at two decimal places. \alg{PCC} performs competitively overall (indeed, the gains of CASSL methods are marginal in many cases), but at a cost of being one of the slowest method in evaluation, on account of more costly inference; (although as mentioned earlier, there are other inference methods exist for PCC, and these may be faster, e.g., \cite{CCAnalysis}). Interestingly, the best performance against PCC was obtained on text datasets. 

Possibly, higher results could be obtained from the \alg{MLP}s with more exhaustive parameter tuning. However, the need for such tuning is an issue in itself; we already tried several different parameter combinations. Furthermore, running time (\Tab{tab:scalability}) is already an order of magnitude (or more) higher than \alg{CCASL} methods due to their iterative learning procedure. An important advantage of the synthetic nodes is eliminating the need for such procedures.

Under Hamming score, \alg{BR$_{RF}$} clearly performs best. This is in line with our earlier discussion, since random forests are powerful non-linear models, and as such are able to ensure that base learners perform well independently under this metric. 
This is the case to some extent with \alg{CCASL} (which also performs very well under this metric), but in some particular cases, \alg{BR$_{RF}$} is more successful.
	
\Fig{fig:contrast} confirms that, while established methods such as \alg{MLP} and \alg{ELM} can also learn difficult concepts (such as the \textsc{xor} label, as widely known in the literature), in practice they need more hidden nodes as compared to our proposed \alg{CCASL} methods. In fact, even a single hidden nodes is enough for \alg{CCASL+AML}. Using PCC-inference (PCCASL) leads to near-perfect performance with small $K$, however, performance actually decreases, or at least becomes more unstable, with larger $K$. The explanation for this is that the number of iterations for MCC is held constant, but the search space increases exponentially with $K$ (namely, $2^{(3+K)}$). More Monte-Carlo iterations will help (indeed, 10 is arguably low), at a computational cost, but the trend is already clear.




	The performance improvement of several of the methods (including \alg{CCASL} and \alg{BR$_{RF}$}) over \alg{BR} and \alg{CC} on \textsf{Medical} is considerable. A closer look revealed that individual label accuracy increases along \alg{CC}'s chain from roughly $0.67$ to $0.75$, and from $0.93$ to $0.99$ in \alg{CCASL}'s chain (excluding the synthetic labels), clearly showing the effect of using synthetic labels first. We already managed to reproduce similar differences on synthetic data (\Tab{fig:srb8}) and, of course, the \textsf{Logical} dataset. 

The gap in running times becomes noticeable under the large datasets, where \alg{CCASL} methods are around twice as expensive as BR/CC. However, if the extra time was used for trialling chain orders with internal train/test split for \alg{CC}, it is probably not enough to make up the gap in predictive performance. \alg{BR$_{RF}$} is also competitively fast. Tree-based methods for multi-label classification also proved themselves in a similar way in \citep{ExtML}.

\begin{table}
	\centering
	\caption{\label{tab:scalability}Average Running Time (seconds) and ranking across $10$ train/test splits.}
	\pink
	\footnotesize
	\begin{tabular}{lrrrrrrrrr}
\hline
Dataset   & \alg{BR} & \alg{CC} & \alg{PCC} & \alg{CCASL} & \alg{C.+BR} & \alg{C.+AML} & \alg{ELM} & \alg{BR$_{RF}$} & \alg{MLP} \\
\hline
Logical    & 0 1 & 0 1 & 0 8 & 0 3 & 0 5 & 0 6 & 0 3 & 0 7 & 1 9  \\ 
Music      & 0 4 & 0 3 & 2 8 & 0 5 & 0 6 & 0 7 & 0 1 & 0 2 & 48 9  \\ 
Scene      & 1 4 & 1 3 & 10 8 & 2 5 & 2 6 & 3 7 & 0 1 & 0 2 & 244 9  \\ 
Yeast      & 0 2 & 0 3 & 33 8 & 1 5 & 2 6 & 3 7 & 0 1 & 1 4 & 114 9  \\ 
Medical    & 5 3 & 5 4 & 31 8 & 12 5 & 12 6 & 13 7 & 0 2 & 0 1 & 243 9  \\ 
Enron      & 19 3 & 20 4 & 85 8 & 37 5 & 38 6 & 41 7 & 1 2 & 1 1 & 965 9  \\ 
Reuters    & 129 3 & 141 4 & 430 8 & 338 6 & 328 5 & 359 7 & 50 2 & 30 1 & 1555 9  \\ 
Ohsumed    & 236 3 & 269 4 & 478 7 & 379 5 & 607 8 & 394 6 & 22 1 & 58 2 & 3081 9  \\ 
M.Mill     & 426 2 & 410 1 & 3753 8 & 904 3 & 1062 6 & 1048 5 & 908 4 & 2295 7 & 5830 9  \\ 
Bibtex     & 1535 3 & 1616 4 & 1934 5 & 3372 6 & 3411 7 & 3440 8 & 141 2 & 60 1 & 6766 9  \\ 
Corel5k    & 464 3 & 566 4 & 2210 9 & 1311 5 & 1423 6 & 1562 7 & 309 2 & 142 1 & 1747 8  \\ 
\hline
avg rank   &  2.86      &  3.23      &  7.73      &  4.86      &  6.09      &  6.73      &  1.95      &  2.64      &  8.91      \\
\hline
\end{tabular}

\end{table}

%



\begin{figure}
	\centering
	\includegraphics[width=0.6\textwidth]{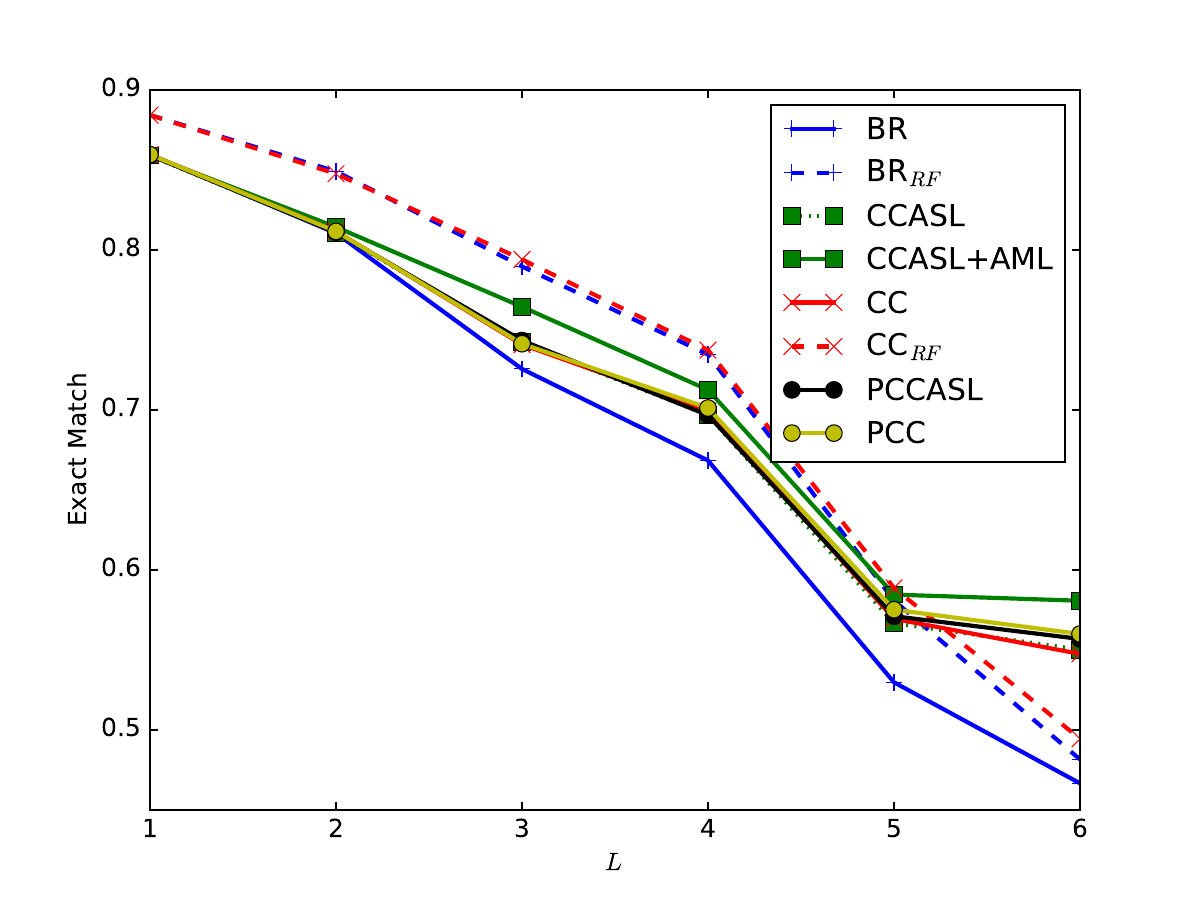}
	\caption{\label{fig:labup} Different methods compared on the \textsf{Scene} dataset for varying numbers of \emph{output labels}. For example, where $L=3$ then only the first $3$ labels of the dataset are considered for learning and evaluation. As in the above experiments $RF$ indicates that random forest was the base learner; in other cases it is logistic regression. As in \Fig{fig:contrast}, PCCASL is CCASL with Monte-Carlo sampling inference.}
\end{figure}

Chaining mechanisms can obtain a greater advantage on datasets with more labels, simply because these labels can be leveraged for a more powerful predictive structure for other labels. \Fig{fig:labup} illustrates this on the \data{Scene} data, where the label variables are incrementally added. For predicting a single label (corresponding to binary classification), random forest clearly outperforms logistic regression. On the other hand, with all six labels, CC-based methods with logistic regression actually performs better, with \alg{CCASL+AML} performing best, followed by others. Notice that the chaining mechanism does not particularly help with random forests -- only a minor improvement of \alg{CC$_{RF}$} over \alg{BR$_{RF}$} is obtained with more labels. The performance signal of \alg{CCASL+AML} separates from other CC-logistic methods at only the second label, whereas the difference between \alg{CC} and \alg{PCC} only becomes noticeable at around $L=5$ or $L=6$. On this particular dataset the performance of CCASL and PCC is virtually indistinguishable, as already observable in \Tab{fig:r2}. Using probabilistic inference (PCCASL) suggests a marginal improvement in this case. The biggest separation in the performance signal is greatest at $L=6$, but does not grow consistently or linearly amoung all methods, indicating again that it is not just the number of labels but also the inherent complexity of each label that affects performance of different methods. 

Our proposed methods showed an advantage in terms of overall ranking and in many particular dataset/metric combinations, however the main goal of the paper was to further investigate and detail the underlying mechanisms responsible for good performance, and in particular the use of labels as hidden nodes. Results indicate the utility of this mechanism. A multi-layer neural network structure is theoretically also as capable as our methods (in terms of approximation ability), however results highlighted the fact that they are difficult to employ in an out-of-the-box fashion. Both approaches use hidden nodes, but it is faster and much easier to use `synthetic labels' than to train a standard back-propagation neural network (MLP), and also much more effective than a random-projection neural network (ELM).
	
We reiterate that it is generally not possible to say that a particular classifier is overall advantageous across all possible scenarios. It was also apparent in our results that different classifiers performed better on certain datasets and under different evaluation metrics: a non-linear decision tree classifier is preferable to logistic regression as a base classifier on \textsf{Ohsumed} and \textsf{MediaMill}, and for several datasets under Hamming loss, but in many other contexts, a linear base learner can be more successfully employed in a classifier-chain or hidden-node classifier. 

In this paper we examined in particular Hamming score and exact match. Dozens of metrics, including for example the micro- and macro-averaged F1 measures, have been used in the literature. A full comparison is outside the scope of this paper, but large-scale empirical comparisons (e.g., \cite{ExtML}) indicate that methods that perform well on both Hamming score and exact match also tend to perform well across a range of other metrics.
	
Perhaps it is also worth remarking that there are now a plethora of multi-label methods, and parameter configurations thereof. Comparing to all of them is outside the scope and possibility of this work, however, future comparisons will help further unravel the mechanisms behind the successful methods, and the contexts wherein this success is most likely to occur.

\section{Summary and Conclusions}
\label{sec:conclusions}


Many methods for multi-label classification measure label dependence in a dataset and then use this information to build structured models with cases of significant dependence manifested explicitly as direct connections among the label outputs. These methods have performed well in the literature in empirical evaluations, but understanding of their mechanism has lagged behind these results. We have contributed to the analysis in this area and looked into this mechanism in greater depth. A common view is that models should be built on a structure that approximates an underlying ground-truth structure in the data. Although in a strict probabilistic sense this may be the case, in practice many shortcuts are taken to deal with scalability concerns, and we explained how in many cases modeling dependence among outputs is simply compensating for an inadequate base classifier or inference scheme, and labels are being leveraged as advanced feature-transformations. We exploited this mechanism in an inner layer, creating efficient methods that perform well even with unconnected labels at the outer layer,   and even with a simple linear base learner, one can still obtain competitive results in this manner. 


We proposed several methods based on the idea of inner `synthetic labels', with different combinations of the labels, both in a cascade form from the features and label subset-composition from the label space. This is not possible in traditional single-label classification where a single target is predicted for each instance. 
Leveraging the synthetic labels means that our methods do not need gradient-based back propagation to train inner nodes: multiple levels are created instantly from combinations of input and output. 
The methods we presented outperformed a variety of related algorithms.  
In future work, we intend to investigate the output of combining this methodology of synthetic labels to speed up existing gradient descent algorithms. 

\section{Acknowledgements}
This work was supported in part by the Aalto Energy Efficiency Research Programme (http://aef.aalto.fi/en/), and Academy of Finland grant 118653 (ALGODAN).

\bibliographystyle{plain}
\bibliography{multilabel.bib,my_publications}

\begin{thebibliography}{10}

\bibitem{PCC}
Krzysztof Dembczy{\'n}ski, Weiwei Cheng, and Eyke H\"ullermeier.
\newblock Bayes optimal multilabel classification via probabilistic classifier
  chains.
\newblock In {\em ICML '10: 27th International Conference on Machine Learning},
  pages 279--286, Haifa, Israel, June 2010. Omnipress.

\bibitem{OnLabelDependence2}
Krzysztof Dembczy\'{n}ski, Willem Waegeman, Weiwei Cheng, and Eyke
  H\"ullermeier.
\newblock On label dependence and loss minimization in multi-label
  classification.
\newblock {\em Mach. Learn.}, 88(1-2):5--45, July 2012.

\bibitem{CCAnalysis}
Krzysztof Dembczy{\'n}ski, Willem Waegeman, and Eyke H\"{u}llermeier.
\newblock An analysis of chaining in multi-label classification.
\newblock In {\em ECAI: European Conference of Artificial Intelligence}, volume
  242, pages 294--299. IOS Press, 2012.

\bibitem{AdaGrad}
John Duchi, Elad Hazan, and Yoram Singer.
\newblock Adaptive subgradient methods for online learning and stochastic
  optimization.
\newblock {\em J. Mach. Learn. Res.}, 12:2121--2159, July 2011.

\bibitem{MECOC}
Chun{-}Sung Ferng and Hsuan{-}Tien Lin.
\newblock Multi-label classification with error-correcting codes.
\newblock In {\em Proceedings of the 3rd Asian Conference on Machine Learning,
  {ACML} 2011, Taoyuan, Taiwan, November 13-15, 2011}, pages 281--295, 2011.

\bibitem{CollectiveMC}
Nadia Ghamrawi and Andrew Mc{C}allum.
\newblock Collective multi-label classification.
\newblock In {\em CIKM '05: 14th ACM international Conference on Information
  and Knowledge Management}, pages 195--200, New York, NY, USA, 2005. ACM
  Press.

\bibitem{MBR}
Shantanu Godbole and Sunita Sarawagi.
\newblock Discriminative methods for multi-labeled classification.
\newblock In {\em PAKDD '04: Eighth Pacific-Asia Conference on Knowledge
  Discovery and Data Mining}, pages 22--30. Springer, 2004.

\bibitem{CDN}
Yuhong Guo and Suicheng Gu.
\newblock Multi-label classification using conditional dependency networks.
\newblock In {\em IJCAI '11: 24th International Conference on Artificial
  Intelligence}, pages 1300--1305. IJCAI/AAAI, 2011.

\bibitem{EoSL}
Trevor Hastie, Robert Tibshirani, and Jerome Friedman.
\newblock {\em The Elements of Statistical Learning}.
\newblock Springer Series in Statistics. Springer New York Inc., New York, NY,
  USA, 2001.

\bibitem{DBNbp}
Geoffrey Hinton and Ruslan Salakhutdinov.
\newblock Reducing the dimensionality of data with neural networks.
\newblock {\em Science}, 313(5786):504 -- 507, 2006.

\bibitem{ELM}
Guang-Bin Huang, DianHui Wang, and Yuan Lan.
\newblock Extreme learning machines: a survey.
\newblock {\em International Journal of Machine Learning and Cybernetics},
  2(2):107--122, 2011.

\bibitem{BeamSearch2}
Abhishek Kumar, Shankar Vembu, AdityaKrishna Menon, and Charles Elkan.
\newblock {Beam search algorithms for multilabel learning}.
\newblock {\em Machine Learning}, 92(1):65--89, 2013.

\bibitem{CRBM}
Hugo Larochelle, Michael Mandel, Razvan Pascanu, and Yoshua Bengio.
\newblock Learning algorithms for the classification restricted {B}oltzmann
  machine.
\newblock {\em J. Mach. Learn. Res.}, 13:643--669, March 2012.

\bibitem{OnTheOptimality}
Weiwei Liu and Ivor Tsang.
\newblock On the optimality of classifier chain for multi-label classification.
\newblock In C.~Cortes, N.~D. Lawrence, D.~D. Lee, M.~Sugiyama, and R.~Garnett,
  editors, {\em Advances in Neural Information Processing Systems 28}, pages
  712--720. Curran Associates, Inc., 2015.

\bibitem{ExtML}
Gjorgji Madjarov, Dragi Kocev, Dejan Gjorgjevikj, and Sa\v{s}o D\v{z}eroski.
\newblock An extensive experimental comparison of methods for multi-label
  learning.
\newblock {\em Pattern Recognition}, 45(9):3084--3104, September 2012.

\bibitem{NN2}
Jinseok Nam, Jungi Kim, Eneldo~Loza Menc{\'{\i}}a, Iryna Gurevych, and Johannes
  F{\"{u}}rnkranz.
\newblock Large-scale multi-label text classification - revisiting neural
  networks.
\newblock In {\em ECML-PKDD '14: 25th European Conference on Machine Learning
  and Knowledge Discovery in Databases}, pages 437--452, 2014.

\bibitem{MEDS2}
Jesse Read, Albert Bifet, Geoff Holmes, and Bernhard Pfahringer.
\newblock Scalable and efficient multi-label classification for evolving data
  streams.
\newblock {\em Machine Learning}, 88(1-2):243--272, 2012.

\bibitem{MCC2}
Jesse Read, Luca Martino, and David Luengo.
\newblock Efficient {M}onte {C}arlo methods for multi-dimensional learning with
  classifier chains.
\newblock {\em Pattern Recognition}, 47(3):1535--1546, 2014.

\bibitem{ECC2}
Jesse Read, Bernhard Pfahringer, Geoffrey Holmes, and Eibe Frank.
\newblock Classifier chains for multi-label classification.
\newblock {\em Machine Learning}, 85(3):333--359, 2011.

\bibitem{SM}
Robert~E. Schapire and Yoram Singer.
\newblock Improved boosting algorithms using confidence-rated predictions.
\newblock {\em Machine Learning}, 37(3):297--336, December 1999.

\bibitem{Dropout}
Nitish Srivastava, Geoffrey Hinton, Alex Krizhevsky, Ilya Sutskever, and Ruslan
  Salakhutdinov.
\newblock Dropout: A simple way to prevent neural networks from overfitting.
\newblock {\em Journal of Machine Learning Research}, 15:1929--1958, 2014.

\bibitem{PLST}
Farbound Tai and Hsuan-Tien Lin.
\newblock Multilabel classification with principal label space transformation.
\newblock {\em Neural Comput.}, 24(9):2508--2542, September 2012.

\bibitem{Tenenboim}
Lena Tenenboim, Lior Rokach, and Bracha Shapira.
\newblock Identification of label dependencies for multi-label classification.
\newblock In {\em MLD '10: 2nd International Workshop on Learning from
  Multi-Label Data from ICML/COLT 2010}, 2010.

\bibitem{Overview}
Grigorios Tsoumakas and Ioannis Katakis.
\newblock Multi label classification: An overview.
\newblock {\em International Journal of Data Warehousing and Mining},
  3(3):1--13, 2007.

\bibitem{RAkEL2}
Grigorios Tsoumakas, Ioannis Katakis, and Ioannis Vlahavas.
\newblock Random k-labelsets for multi-label classification.
\newblock {\em IEEE Transactions on Knowledge and Data Engineering},
  23(7):1079--1089, 2011.

\bibitem{HOMER}
Grigorios Tsoumakas, Ioannis Katakis, and Ioannis~P. Vlahavas.
\newblock Effective and efficient multilabel classification in domains with
  large number of labels.
\newblock In {\em ECML/PKDD Workshop on Mining Multidimensional Data}, 2008.

\bibitem{CMAC}
{W. Thomas Miller III}, Filson~H. Glanz, and {L. Gordon Kraft III}.
\newblock {CMAC}: An associative neural network alternative to backpropagation.
\newblock {\em Proceedings of the {IEEE}}, 78(10):1561--1567, October 1990.

\bibitem{KDE}
Jason Weston, Olivier Chapelle, Andr{\'e} Elisseeff, Bernhard Sch{\"o}lkopf,
  and Vladimir Vapnik.
\newblock Kernel dependency estimation.
\newblock In {\em Advances in Neural Information Processing Systems 15 (NIPS)},
  pages 897--904, 2003.

\bibitem{BCC}
Julio~H. Zaragoza, Luis~Enrique Sucar, Eduardo~F. Morales, Concha Bielza, and
  Pedro Larra{\~n}aga.
\newblock Bayesian chain classifiers for multidimensional classification.
\newblock In {\em 24th International Joint Conference on Artificial
  Intelligence (IJCAI '11)}, pages 2192--2197, 2011.

\bibitem{MLRBF}
Min-Ling Zhang.
\newblock {ML-RBF} : {RBF} neural networks for multi-label learning.
\newblock {\em Neural Processing Letters}, 29(2):61--74, 2009.

\bibitem{LIFT2}
Min-Ling Zhang and Lei Wu.
\newblock Lift: Multi-label learning with label-specific features.
\newblock {\em Pattern Analysis and Machine Intelligence, IEEE Transactions
  on}, 37(1):107--120, Jan 2015.

\bibitem{LEAD}
Min-Ling Zhang and Kun Zhang.
\newblock Multi-label learning by exploiting label dependency.
\newblock In {\em KDD '10: 16th ACM SIGKDD International conference on
  Knowledge Discovery and Data mining}, pages 999--1008. ACM, 2010.

\bibitem{BPMLL}
Min-Ling Zhang and Zhi-Hua Zhou.
\newblock Multilabel neural networks with applications to functional genomics
  and text categorization.
\newblock {\em IEEE Transactions on Knowledge and Data Engineering},
  18(10):1338--1351, 2006.

\end{thebibliography}

\end{document}